\documentclass{article}
\usepackage{log_2022}         

\usepackage{booktabs}           
\usepackage{multirow}           
\usepackage{amsfonts}           
\usepackage{graphicx}           
\usepackage{duckuments}         

\usepackage[numbers,compress,sort]{natbib}

\usepackage{url}            
\usepackage{booktabs}       
\usepackage{amsfonts}       
\usepackage{nicefrac}       
\usepackage{microtype}      
\usepackage{xcolor}         
\usepackage{natbib}
\usepackage{tikz}
\usepackage{microtype}
\usepackage{graphicx}
\usepackage{booktabs} 
\usepackage{comment}
\usepackage{enumitem}
\usepackage{amsthm}
\usepackage{float}
\usepackage{bbm}
\usepackage{amsmath}
\usepackage{mathtools}
\usepackage{multirow}
\usepackage{diagbox}
\usepackage{comment}
\usepackage{thmtools,thm-restate}

\usepackage{rotating}
\usepackage{bm}
\usepackage{subfig}
\usepackage{booktabs, makecell, tabularx}
\usepackage{rotating}
\usepackage[export]{adjustbox}
\usepackage{algorithm}
\let\classAND\AND
\let\AND\relax
\usepackage{algorithmic}

\let\AND\classAND
\AtBeginEnvironment{algorithmic}{\let\AND\algoAND}

\floatname{algorithm}{Algorithm}

\usepackage{xcolor}         
\definecolor{CColor}{rgb}{0.01,0.31,0.59}
\definecolor{GGray}{rgb}{0.80,0.90,1}
\definecolor{Shady}{rgb}{0.9,0.9,0.9}
\definecolor{kaistblue}{RGB}{20,135,200}
\definecolor{kaistdarkblue}{RGB}{0,65,145}
\definecolor{urbanablue}{RGB}{19,41,75}
\definecolor{urbanaorange}{RGB}{232,74,39}
\definecolor{drp}{rgb}{0.53,0.15,0.34}

\newcommand*\circled[1]{\tikz[baseline=(char.base)]{\node[shape=circle,draw,inner sep=0.4pt] (char) {#1};}}

\makeatletter
\def\thickhline{%
  \noalign{\ifnum0=`}\fi\hrule \@height \thickarrayrulewidth \futurelet
   \reserved@a\@xthickhline}
\def\@xthickhline{\ifx\reserved@a\thickhline
               \vskip\doublerulesep
               \vskip-\thickarrayrulewidth
             \fi
      \ifnum0=`{\fi}}
\makeatother

\newlength{\thickarrayrulewidth}
\title[Better Graph Neural Networks by Not Training Weights at All: Finding Untrained GNNs Tickets]{You Can Have Better Graph Neural Networks by Not Training Weights at All: Finding Untrained GNNs Tickets}

\author[T. Huang et al.]{%
  Tianjin Huang\textsuperscript{1},
  Tianlong Chen\textsuperscript{2}, 
  Meng Fang\textsuperscript{1,4},
  Vlado Menkovski\textsuperscript{1},\\
  \textbf{Jiaxu Zhao\textsuperscript{1},}
  \textbf{Lu Yin\textsuperscript{1},}
  \textbf{Yulong Pei\textsuperscript{1},}
  \textbf{Decebal Constantin Mocanu\textsuperscript{1,3},}\\
  \textbf{Zhangyang Wang\textsuperscript{2},}
  \textbf{Mykola Pechenizkiy\textsuperscript{1},}
  \textbf{Shiwei Liu\textsuperscript{2}}
  \\
  {\textsuperscript{1}Eindhoven University of Technology, \textsuperscript{2}University of Texas at Austin}\\
  {\textsuperscript{3}University of Twente}, 
  {\textsuperscript{4}University of Liverpool}\\
    \small{\texttt{\{t.huang,j.zhao,v.menkovski,l.yin,y.pei.1,m.pechenizkiy\}@tue.nl}}\\
  \small{\texttt{\{tianlong.chen,atlaswang\}@utexas.edu}} \\
  \small{\texttt{\{d.c.mocanu\}@utwente.nl}},
  \small{\texttt{\{Meng.Fang\}@liverpool.ac.uk}} \\
  \small{\texttt{shiwei.liu@austin.utexas.edu}}
  \\
}

\begin{document}
\maketitle

\begin{abstract}
Recent works have impressively demonstrated that there exists a subnetwork in randomly initialized convolutional neural networks (CNNs) that can match the performance of the fully trained dense networks at initialization, without any optimization of the weights of the network (i.e., untrained networks). However, the presence of such untrained subnetworks in graph neural networks (GNNs) still remains mysterious. In this paper we carry out the first-of-its-kind exploration of discovering matching untrained GNNs. With sparsity as the core tool, we can find \textit{untrained sparse subnetworks} at the initialization, that can match the performance of \textit{fully trained dense} GNNs. Besides this already encouraging finding of comparable performance, we show that the found untrained subnetworks can substantially mitigate the GNN over-smoothing problem, hence becoming a powerful tool to enable deeper GNNs without bells and whistles. We also observe that such sparse untrained subnetworks have appealing performance in out-of-distribution detection and robustness of input perturbations. We evaluate our method across widely-used GNN architectures on various popular datasets including the Open Graph Benchmark (OGB). The code is available at \url{https://github.com/TienjinHuang/UGTs-LoG.git}. 
\end{abstract}

\section{Introduction}\label{intro}

Graph Neural Networks (GNNs)~\citep{scarselli2008graph,li2015gated} have shown the power to learn representations from graph-structured data. Over the past decade, GNNs and their variants such as Graph Convolutional Networks (GCN)~\citep{kipf2016semi}, Graph Isomorphism Networks (GIN)~\citep{xu2018how}, Graph Attention Networks (GAT)~\citep{velivckovic2017graph} have been successfully applied to a wide range of scenarios, e.g., social analysis~\citep{qiu2018deepinf,li2019predicting}, protein feature learning~\citep{zitnik2017predicting}, traffic prediction~\citep{guo2019attention}, and recommendation systems~\citep{ying2018graph}. In parallel, works on untrained networks~\citep{zhou2019deconstructing,ramanujan2020s} surprisingly discover the presence of untrained subnetworks in CNNs that can already match the accuracy of their fully trained dense CNNs with their initial weights, without any weight update. In this paper, we attempt to explore discovering untrained sparse networks in GNNs by asking the following question: 

\vspace{0.1em}
\textit{Is it possible to find a well-performing graph neural (sub-) network without any training of the model weights?} 
\vspace{0.1em}

Positive answers to this question will have significant impacts on the research field of GNNs.~\textbf{\circled{1}} If the answer is yes, it will shed light on a new direction of obtaining performant GNNs, e.g., traditional training might not be indispensable towards performant GNNs.
~\textbf{\circled{2}}  The existence of such performant subnetworks will extend the recently proposed untrained subnetwork techniques~\citep{zhou2019deconstructing,ramanujan2020s} in GNNs. Prior works~\citep{zhou2019deconstructing,ramanujan2020s,chijiwa2021pruning} successfully find that randomly weighted full networks contain untrained subnetworks which perform well without ever modifying the weights, in convolutional neural networks (CNNs). However, the similar study has never been discussed for GNNs. While CNNs reasonably contain well-performing untrained subnetworks due to heavy over-parameterization, GNN models are usually much more compact, and it is unclear whether a performant subnetwork ``should'' still exist in GNNs. 

Furthermore, we investigate the connection between untrained sparse networks and widely-known barriers in deep GNNs, such as over-smoothing. For instance, as analyzed in~\cite{li2018deeper}, by naively stacking many layers and adding non-linearity, the output features are prone to collapsing and becoming indistinguishable. Such undesirable properties significantly limit the power of deeper/wider GNNs, hindering the potential application of GNNs on large-scale graph datasets such as the latest Open Graph Benchmark (OGB)~\citep{hu2020open}. It is interesting to see what would happen for untrained graph neural networks. Note that the goal of sparsity in our paper is \textbf{not for efficiency}, but to obtain nontrivial predictive performance without training (a.k.a., ``masking is training'' \cite{zhou2019deconstructing}). We summarize our contributions as follows:

\begin{figure}[t]
\centering
\subfloat{\includegraphics[width=1.0\textwidth]{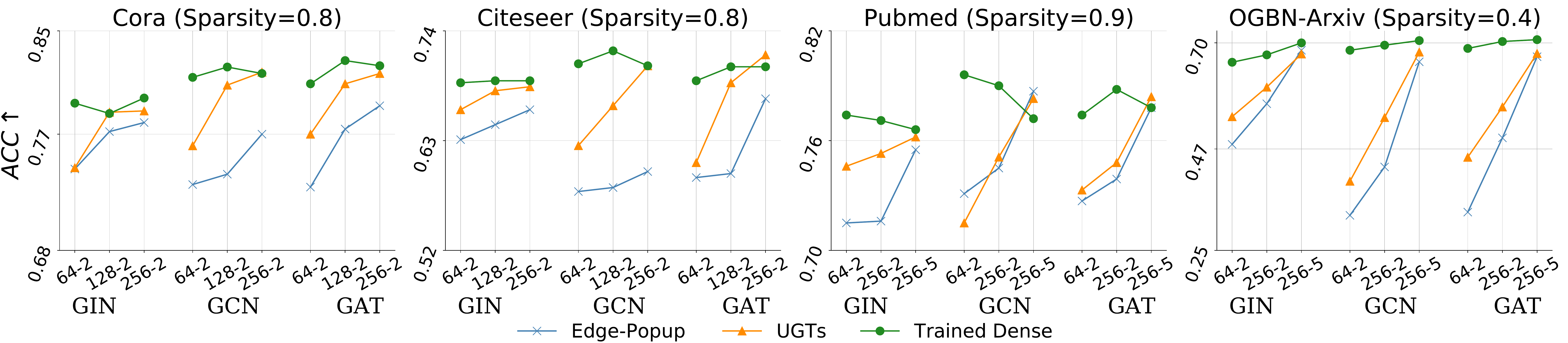} 
}
\caption{
Performance of untrained graph subnetworks (UGTs (ours) and Edge-Popup~\citep{ramanujan2020s}) and the corresponding trained dense GNNs. We demonstrate that as the model size increases, UGTs is able to find an untrained subnetwork with its random initializations, that can match the performance of the corresponding fully-trained dense GNNs. The x-axis denotes the corresponding model size for each point, e.g. ``64-2'' represents a model with 2 layers and width 64.}  
\label{fig:Effectiveness}
\end{figure}

\begin{itemize}
    \item We demonstrate for the first time that there exist untrained graph subnetworks with matching performance (referring to as good as the trained full networks),~\textbf{within randomly initialized dense networks and without any model weight training}. Distinct from the popular lottery ticket hypothesis (LTH)~\citep{frankle2018lottery,chen2021unified}, neither the original dense networks nor the identified subnetworks need to be trained.  
    
    \item We find that the gradual sparsification  technique~\citep{zhu2017prune,gale2019state} can be a stronger performance booster. Leveraging its global sparse variant~\citep{liu2021sparse}, we propose our method -- UGTs, which discovers matching untrained subnetworks within the dense GNNs at~\underline{extremely high sparsities}. For example, our method discovers untrained matching subnetworks with up to 99\% sparsity. We validate it across various GNN architectures (GCN, GIN, GAT) on eight datasets, including the large-scale  OGBN-ArXiv and OGBN-Products.
    
    \item We empirically show a surprising observation that our method significantly mitigates the over-smoothing problem without any additional tricks and can successfully scale GNNs up with negligible performance loss. Additionally, we show that UGTs also enjoys favorable performance on Out-of-Distribution (OOD) detection and robustness on different types of perturbations.
\end{itemize}

\section{Related Work}
\textbf{Graph Neural Networks.} 
Graph neural networks is a powerful deep learning approach for graph-structured data. Since proposed in~\citep{scarselli2008graph}, many variants of GNNs have been developed, e.g., GAT~\citep{velivckovic2017graph}, GCN~\citep{kipf2016semi}, GIN~\citep{xu2018how}, GraphSage~\citep{hamilton2017inductive}, SGC~\citep{wu2019simplifying}, and GAE~\citep{kipf2016variational}. More and more recent works point out that deeper GNN architectures potentially provide benefits to practical graph structures, e.g., molecules~\citep{zitnik2017predicting}, point clouds~\citep{li2019deepgcns}, and
meshes~\citep{gong2020geometrically}, as well as large-scale graph dataset OGB. However, training deep GNNs usually is a well-known challenge due to various difficulties such as gradient vanishing and over-smoothing problems~\citep{li2018deeper,zhao2019pairnorm}. The existing approaches to address the above-mentioned problem can be categorized into three groups: (1) skip connection, e.g., Jumping connections~\citep{xu2018representation,liu2020towards}, Residual connections~\citep{li2019deepgcns}, and Initial connections~\citep{chen2020simple}; (2) graph normalization, e.g., PairNorm~\citep{zhao2019pairnorm}, NodeNorm~\citep{zhou2020understanding}; (3) random dropping including DropNode~\citep{huang2020tackling} and DropEdge~\citep{rong2019dropedge}.


\textbf{Untrained Subnetworks.} 
Untrained subnetworks refer to the hypothesis that there exists a subnetwork in a randomly intialized neural network that can achieve almost the same accuracy as a fully trained neural network without weight update. 
\cite{zhou2019deconstructing} and~\cite{ramanujan2020s} first demonstrate that randomly initialized CNNs contain subnetworks that achieve impressive performance without updating weights at all.~\cite{chijiwa2021pruning} enhanced the performance of untrained subnetworks by iteratively reinitializing the weights that have been pruned. Besides the image classification task, some works also explore the power of untrained subnetworks in other domains, such as multi-tasks learning~\cite{wortsman2020supermasks} and adversarial robustness~\cite{fu2021drawing}. 

Instead of proposing well-versed techniques to enable deep GNNs training, we explore the possibility of finding well-performing deeper graph subnetworks at initialization in the hope of avoiding the difficulties of building deep GNNs without model weight training.

\section{Untrained GNNs Tickets}
\label{sec:pre_setup}
\subsection{Preliminaries and Setups}
\textbf{Notations.} We represent matrices by bold uppercase characters, e.g. $\bm{X}$, vectors by bold lowercase characters, e.g. $\bm{x}$, and scalars by normal lowercase characters, e.g. x. 
We denote the $i^{th}$ row of a matrix $\bm{A}$ by  $\bm{A}[i,:]$,  and  the $(i,j)^{th}$ element of matrix $\bm{A}$ by $\bm{A}[i,j]$. 
 We consider a graph $\mathcal{G}=\{\mathcal{V},\mathcal{E}\}$ where $\mathcal{E}$ is a set of edges and $\mathcal{V}$ is a set of nodes. Let $g(\bm{A},\bm{X};\bm{\theta})$ be a graph neural network where $\bm{A}\in \{0,1\}^{|V|\times |V|}$ is adjacency matrix for describing the overall graph topology, and $\bm{X}$ denotes nodal features . $\bm{A}[i,j]=1$ denotes the edge between node $v_{i}$ and node $v_{j}$. Let $f(\bm{X};\bm{\theta})$ be a neural network with the weights $\bm{\theta}$. $\|\cdot\|_{0}$ denotes the $L_{0}$ norm. 
 
\textbf{Sparse Neural Networks.}  Given a dense network $\bm{\theta_l} \in \mathcal{R}^{d_l}$ with a dimension of $d_{l}$ in each layer $l \in \{1,...,L\}$, binary mask $\bm{m_{l}}\in \{0,1\}^{d_l}$ yielding a sparse Neural Networks with sparse weights $\bm{\theta_l}\odot \bm{m_l}$. The sparsity level is the fraction of the weights that are zero-valued, calculated as $s=1-\frac{\sum_{l}{\|\bm{m_l}\|_{0}}}{\sum_{l}{d_l}}$.   

\textbf{Graph Neural Networks.} GNNs denote a family of algorithms that extract structural information from graphs~\citep{hamilton2017representation} and it is consisted of \emph{Aggregate} and \emph{Combine} operations.
Usually, \emph{Aggregate} is a function that aggregates messages from its neighbor nodes, and \emph{Combine} is an update function that updates the representation of the current node.  Formally, given the graph $\mathcal{G}=(\bm{A},\bm{X})$ with node set $\mathcal{V}$ and edge set $\mathcal{E}$, the $l$-th layer of a GNN is represented as follows:
\begin{equation}
    \bm{a}_{v}^{l}=Aggregate^{l}(\{\bm{h}_{u}^{l-1} : \forall u \in \mathcal{N}(v)\})
\end{equation}
\begin{equation}
    \bm{h}_{v}^{l} = Combine^{l}(\bm{h}_{v}^{l-1},\bm{a}_{v}^{l})
\end{equation}
where $\bm{a}_{v}^{l}$ is the aggregated representation of the neighborhood for node $v$ and $\mathcal{N}(v)$ denotes the neighbor nodes set of the node $v$, and $\bm{h}_{v}^{l}$ is the node representations at the $l$-th layer. After propagating through $L$ layers, we achieve the final node representations $\bm{h}_{v}^{L}$ which can be applied to downstream node-level tasks, such as node classification, link prediction.


\textbf{Untrained Subnetworks.} Following the prior work~\citep{zhou2019deconstructing},~\cite{ramanujan2020s} proposed Edge-Popup which enables finding untrained subnetworks hidden in the 
a randomly initialized full network $f(\bm{\theta})$ by solving the following discrete optimization problem: 
\begin{equation}
    \min_{\bm{m}(\bm{S}) \in \{0,1\}^{|\bm{\theta}|}} \mathcal{L}(f(\bm{A},\bm{X}; \bm{\theta} \odot \bm{m}(\bm{S})),\bm{y})
\end{equation}
where $\mathcal{L}$ is task-dependent loss function; $\odot$ represents an element-wise multiplication; $\bm{y}$ is the label for the input $\bm{X}$ and $\bm{m}$ is the binary mask that controls the sparsity level $s$. $\bm{S}$ is the latent score behind the binary mask $\bm{m}$ and it has the same dimension as $\bm{m}$. To avoid confusion, here we use $\bm{m}(\bm{S})$ instead of $\bm{m}$ to indicate that $\bm{m}$ is generated by $\bm{S}$. We will use $\bm{m}$ directly for brevity in the following content. 

Different from the traditional training of deep neural networks, here the network weights are never updated, masks $\bm{m}$ are instead generated to search for the optimal untrained subnetwork. In practice, each mask $\bm{m}_{i}$ has a latent score variable $\bm{S}_{i} \in \mathcal{R}$ that represents the importance score of the corresponding weight $\bm{\theta}_{i}$.
During training in the forward pass, the binary mask $\bm{m}$ is generated by setting \underline{top-${s}$} smallest elements of $\bm{S}$ to 0 otherwise 1. In the backward pass, all the values in $\bm{S}$ will be updated with straight-through estimation~\citep{bengio2013estimating}. At the end of the training, an untrained subnetwork can be found by the generated mask $\bm{m}$ according to the converged scores $\bm{S}$.

\begin{figure*}[h]
\centering
\subfloat{\includegraphics[width=0.98\textwidth]{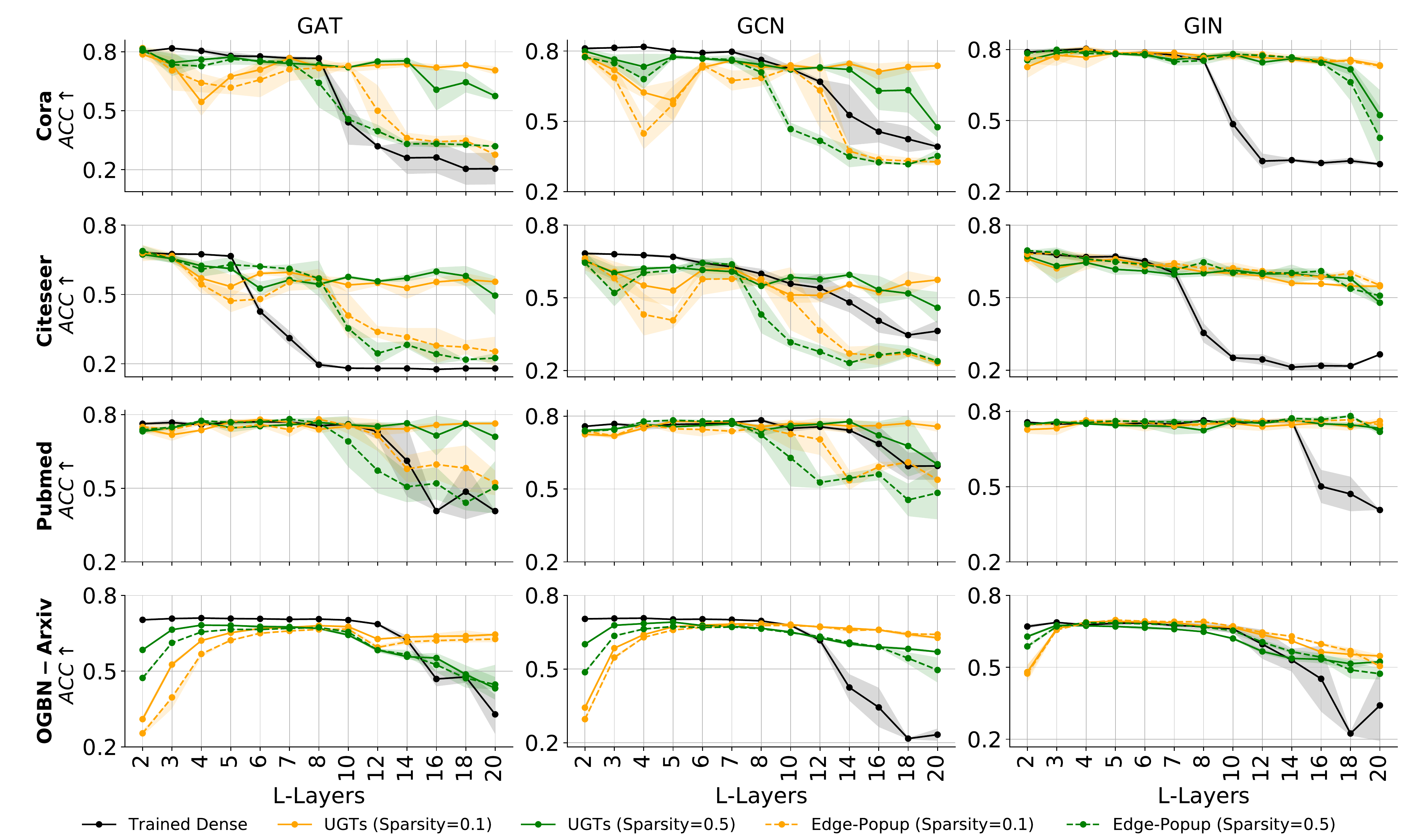}}
\caption{The performance of GNNs  with increasing model depths.  Experiments are conducted on various GNNs with Cora, Citeseer, Pubmed and OGBN-Arxiv. We observe that as the model goes deeper, fully-trained dense GNNs suffer from a sharp accuracy drop, while UGTs preserves the high accuracy. All the results reported are averaged from 5 runs.}

\label{fig:layers}
\end{figure*}

\subsection{Untrained GNNs Tickets -- UGTs}

In this section, we adopt the untrained subnetwork techniques to GNNs and introduce our new approach -- Untrained GNNs Tickets (UGTs).   We share the pseudocode of UGTs in the Appendix~\ref{apped_psu}.


Formally, given a graph neural network $g(\bm{A},\bm{X};\bm{\theta})$, where $\bm{A}$ and $\bm{X}$ are adjacency matrix and nodal features respectively. The optimization problem of finding an untrained subnetwork in GNNs can be therefore described as follows:
\begin{align}
    \min_{\bm{m}\in \{0,1\}^{|\bm{\theta}|}} \mathcal{L}(g(\bm{A},\bm{X};\bm{\theta} \odot \bm{m}),\bm{y})     \label{eq:optimization}
\end{align}
Although Edge-Popup~\citep{ramanujan2020s} can find untrained subnetworks with proper predictive accuracy, its performance is still away from satisfactory. For instance,  Edge-Popup can only obtain matching subnetworks at a relatively low sparsity i.e., 50\%. 

We highlight two limitations of the existing prior research. First of all, prior works~\citep{ramanujan2020s,chijiwa2021pruning} \textbf{initially} set the sparsity level of $\bm{m}_{i}$  as $s$ and maintain it throughout the optimization process. This is very appealing for the scenarios of sparse training~\citep{mocanu2018scalable,onemillionneurons,evci2020rigging} that chases a better trade-off between performance and efficiency, since the fixed sparsity usually translates to fewer floating-point operations (FLOPs). This scheme, however, is not necessary and perhaps harmful to the finding of the smallest possible untrained subnetwork that still performs well. Particularly as shown in~\cite{liu2021sparse}, larger searching space for sparse neural networks at the early optimization phase leads to better sparse solutions. The second limitation is that the existing methods sparsify networks layer-wise with a uniform sparsity ratio, which typically leads to inferior performance compared with the non-uniform layer-wise sparsity~\citep{evci2020rigging,kusupati2020soft,liu2021sparse}, especially for deep architectures~\citep{dettmers2019sparse}. 

\begin{table*}[!htb]
\centering
\caption{Test accuracy (\%) of different training techniques. The experiments are based on GCN models with 16, 32 layers, respectively. Width is set to  448. See Appendix~\ref{appedix_oversmoothing} for GAT architecture. The results of the other methods are obtained from~\cite{chen2021bag}. }
\begin{adjustbox}{width=0.8\textwidth}
\begin{tabular}{c |cc|cc|cc}
\toprule[2pt]
   &\multicolumn{2}{c|}{Cora} &\multicolumn{2}{c|}{Citeseer}&\multicolumn{2}{c}{Pubmed} \\
\midrule[1pt]
    N-Layers&   16 & 32 &   16 & 32 &  16 & 32  \\
\midrule[2pt]
    \textbf{Trained Dense GCN} &21.4 &21.2  &19.5 &20.2  &39.1 &38.7\\
\midrule[1pt]
    +Residual  &20.1 & 19.6&20.8 &20.90  &38.8&38.7 \\
     +Jumping &76.0 &75.5 & 58.3&55.0 &75.6 &75.3  \\
\midrule[1pt]
    +NodeNorm &21.5 &21.4 &18.8&19.1 &18.9&18 \\
    +PairNorm &55.7 &17.7  &27.4 &20.6  &71.3 &61.5\\
\midrule[1pt]
    +DropNode &27.6 &27.6&21.8 &22.1 &40.3 &40.3 \\
    +DropEdge&28.0&27.8&22.9&22.9&40.6 &40.5 \\
\midrule[1.5pt]
    \textbf{UGTs-GCN}& \textbf{77.3 $\pm$ 0.9} &\textbf{77.5 $\pm$ 0.8}  &\textbf{61.1$\pm$0.9} &\textbf{56.2$\pm$0.4} &\textbf{77.6$\pm$0.9} &\textbf{76.3$\pm$1.2}\\
\bottomrule[2pt]
\end{tabular}
\end{adjustbox}
\label{tab:effective_oversmooth}
\end{table*}

\textbf{Untrained GNNs Tickets (UGTs)}. Leveraging the above-mentioned insights, we propose a new approach UGTs here which can discover matching untrained subnetworks with extremely high sparsity levels, i.e., up to 99\%. Instead of keeping the sparsity of $\bm{m}$ fixed throughout the sparsification process, we start from an untrained dense GNNs  and gradually increase the sparsity to the target sparsity during the whole sparsification process. We adjust the original gradual sparsification schedule ~\citep{zhu2017prune,gale2019state} to the linear decay schedule, since no big performance difference can be observed. The 
sparsity level $s_{t}$ of each adjusting step $t$ is calculated as follows:
\begin{align}
    s_{t}=s_{f}+(s_{i}-s_{f})(1-\frac{t-t_{0}}{n \Delta t})  \label{gradual_decay} \\ 
    t\in \{t_{0},t_{0}+\Delta t,...,t_{0}+n\Delta t\} \notag
\end{align}
where $s_{f}$ and $s_{i}$ refer to the final sparsity and initial sparsity, respectively; The initial sparsity is the sparsity at the start point of sparsification and it is set to $0$ in this study. The final sparsity is the sparsity at the endpoint of sparsification.  $t_0$ is the starting point of sparsification; $\Delta t$ is the time between two adjusting steps; $n$ is the total number of adjusting steps.  We set $\Delta t$ as one epoch of mask optimization in this paper. 

To obtain a good non-uniform layer-wise sparsity ratio, we remove the weights with the smallest score values ($\bm{S}$) across layers at each adjusting step. We do this  because~\cite{liu2021sparse} showed that the layer-wise sparsity obtained by this scheme outperforms the other well-studied sparsity ratios~\citep{gale2019state,mocanu2018scalable,evci2020rigging}. More importantly, removing weights across layers theoretically has a larger search space than solely considering one layer. The former can be more appealing as the GNN architecture goes deeper.



\section{Experimental Results}\label{exp}
In this section, we conduct extensive experiments among multiple GNN architectures and datasets to evaluate UGTs. We summarize the experimental setups here. 

\begin{table}[!h]
\vspace{-0.1in}
\centering
\caption{Graph datasets statistics.}\label{datasets}
\begin{adjustbox}{width=0.9\textwidth}
\begin{tabular}{l|c|c|c|c|c|c}
\toprule[2pt]
DataSets &\#Graphs &\#Nodes &\#Edges &\#Classes &\#Features&Metric \\
\midrule[2pt]
Cora  &1& 2708&5429 &7&1433 &Accuracy\\
Citeseer&1 & 3327&4732 &6&3703&Accuracy\\
Pubmed&1 &19717&44338 &3 &3288&Accuracy\\
OGBN-Arxiv&1 &169343&1166243&40&128&Accuracy\\
Texas&1 &183&309&5&1703&Accuracy\\
OGBN-Products &1 &24449029&61859140&47&100&Accuracy\\ 
OGBG-molhiv &41127&25.5(Average)&27.5(Average)&2&-&ROC-AUC\\
OGBG-molbace&1513&34.1(Average)&36.9(Average)&2&-&ROC-AUC\\
\bottomrule[2pt]
\end{tabular}
\end{adjustbox}
\vspace{-0.1in}
\label{tab:dataDescription}
\end{table}

\textbf{GNN Architectures}. We use the three most widely used GNN architectures: GCN, GIN, and GAT~\footnote{All experiments based on GAT architecture are conducted with heads=1 in this study.} in our paper.

\textbf{Datasets}. We choose three popular small-scale graph datasets including Cora, Citeseer, PubMed~\citep{kipf2016semi} and one latest large-scale graph dataset OGBN-Arxiv~\citep{hu2020open} for our main experiments. To draw a solid conclusion, we also evaluate our method on other datasets including  OGBN-Products ~\citep{hu2020open}, TEXAS~\cite{pei2020geom}, OGBG-molhiv~\cite{hu2020open} and OGBG-molbace~\cite{hu2020open,wu2018moleculenet}.  
More detailed information can be found in Table~\ref{tab:dataDescription}.

\subsection{The Existence of Matching Subnetworks}
Figure~\ref{fig:Effectiveness} shows the effectiveness of UGTs with different GNNs, including GCN, GIN and GAT, on the four datasets.
We can observe that as the model size increases, UGTs can find untrained subnetworks that match the fully-trained dense GNNs. This observation is perfectly in line with the previous findings~\citep{ramanujan2020s,chijiwa2021pruning}, which reveal that model size plays a crucial role to the existence of matching untrained subnetworks. Besides, it can be observed that the proposed UGTs consistently outperforms Edge-Popup across different settings.

\begin{figure*}[htb!]
\centering
\setlength\tabcolsep{0.4pt}
\settowidth\rotheadsize{Radcliffe Cam}
\begin{adjustbox}{width=0.91\textwidth}
\begin{tabularx}{1\linewidth}{c ccccc ccccc}
\midrule[2pt]
&Cora&Citeseer&Pubmed &Cora & Citeseer&Pubmed \\
\midrule[2pt]
\rothead{\centering{Trained Dense-16}} 
                        &   \includegraphics[width=0.15\textwidth,valign=m]{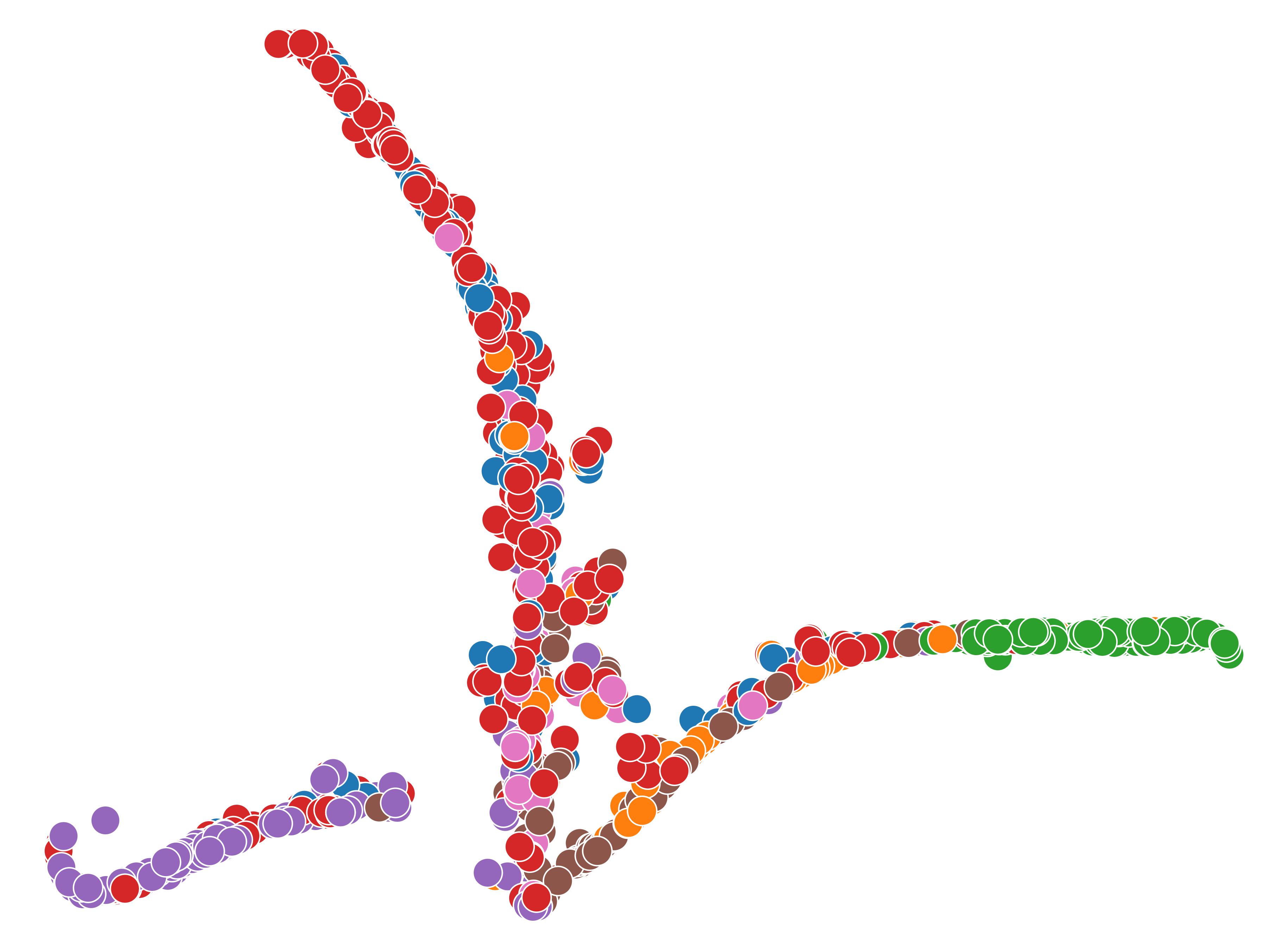}    
                        &   \includegraphics[width=0.15\textwidth,valign=m]{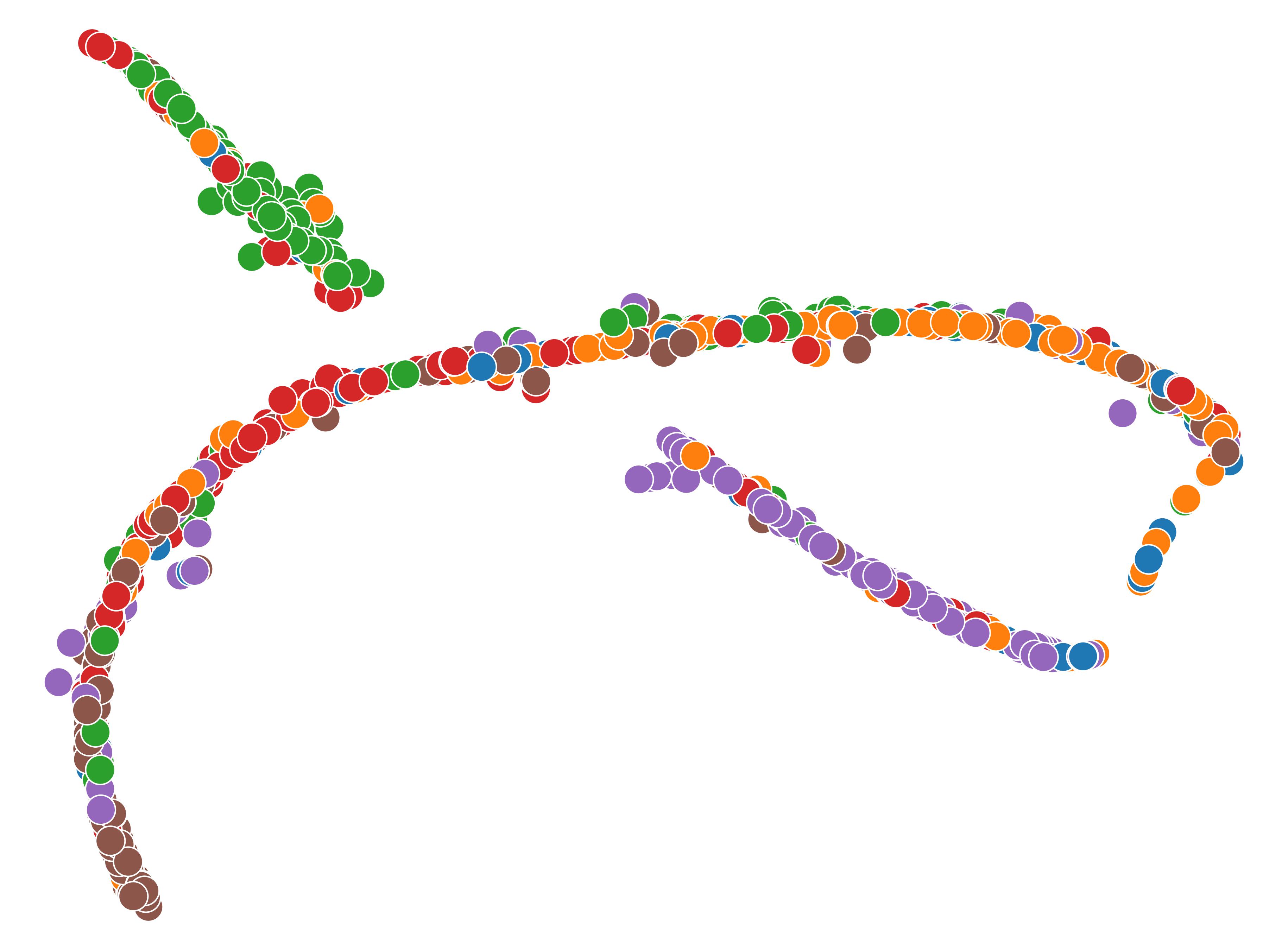}
                        &   \includegraphics[width=0.15\textwidth,valign=m]{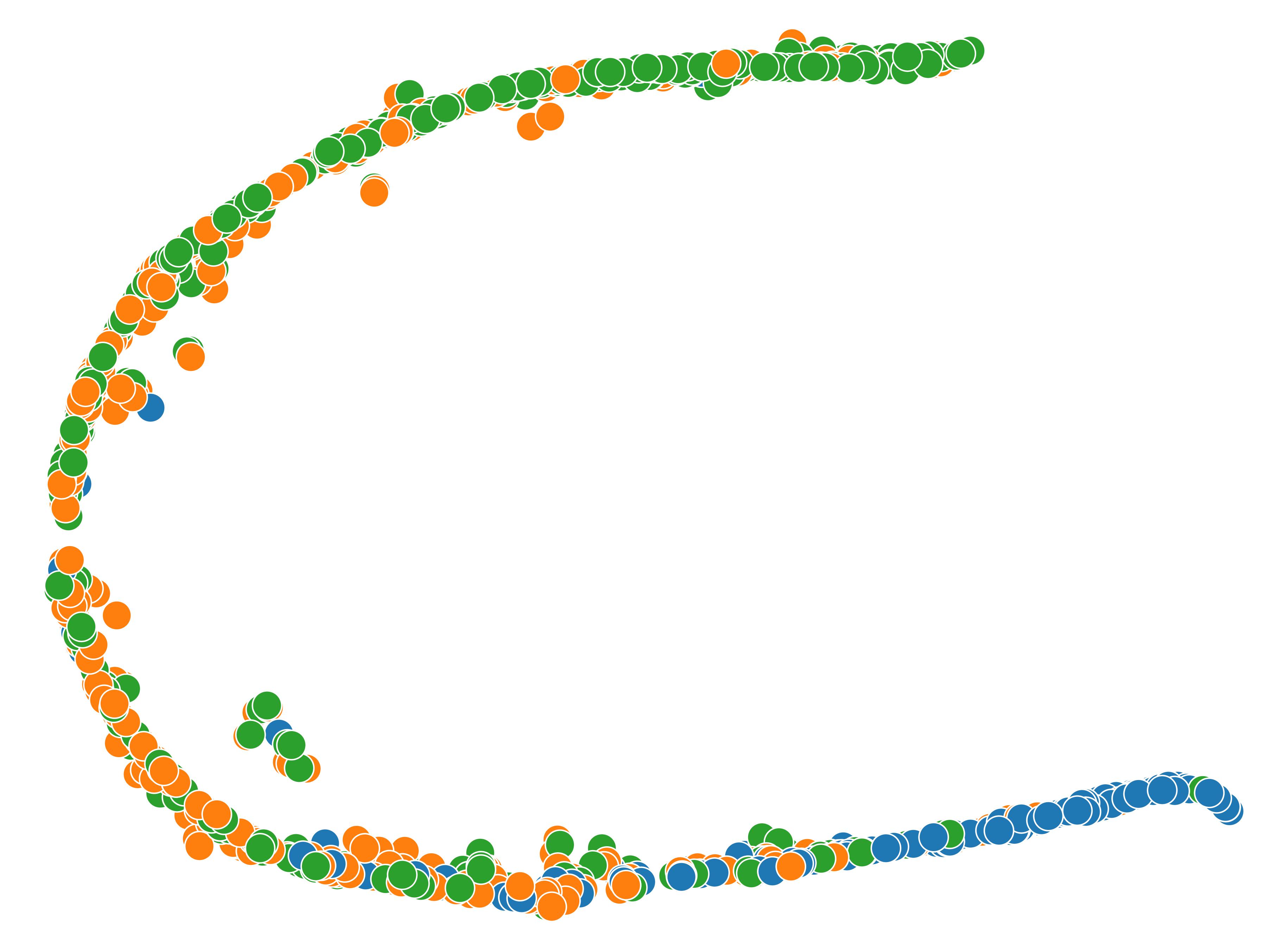} 
\rothead{\centering{Trained Dense-32}} 
                        &   \includegraphics[width=0.15\textwidth,valign=m]{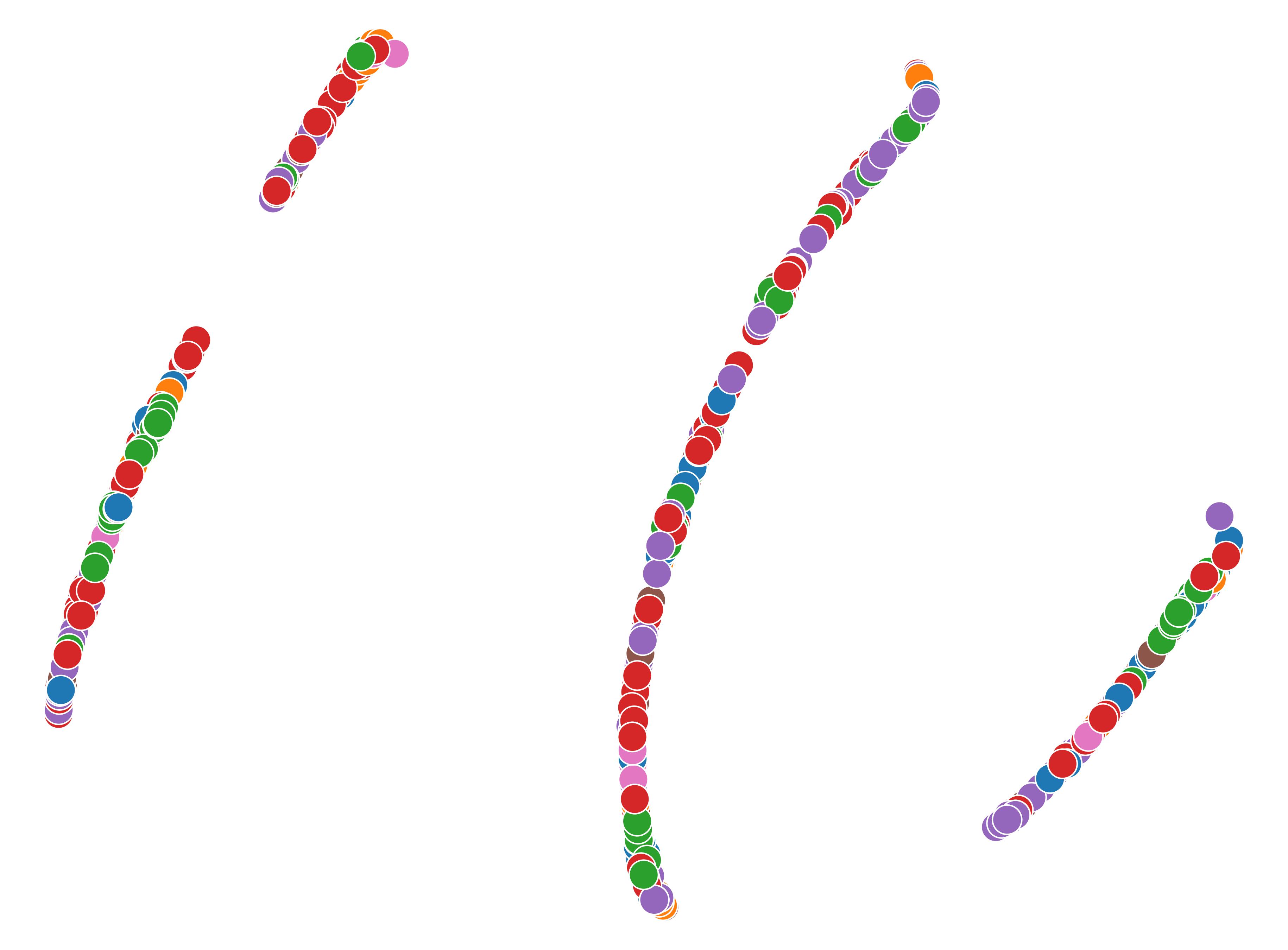}    
                        &   \includegraphics[width=0.15\textwidth,valign=m]{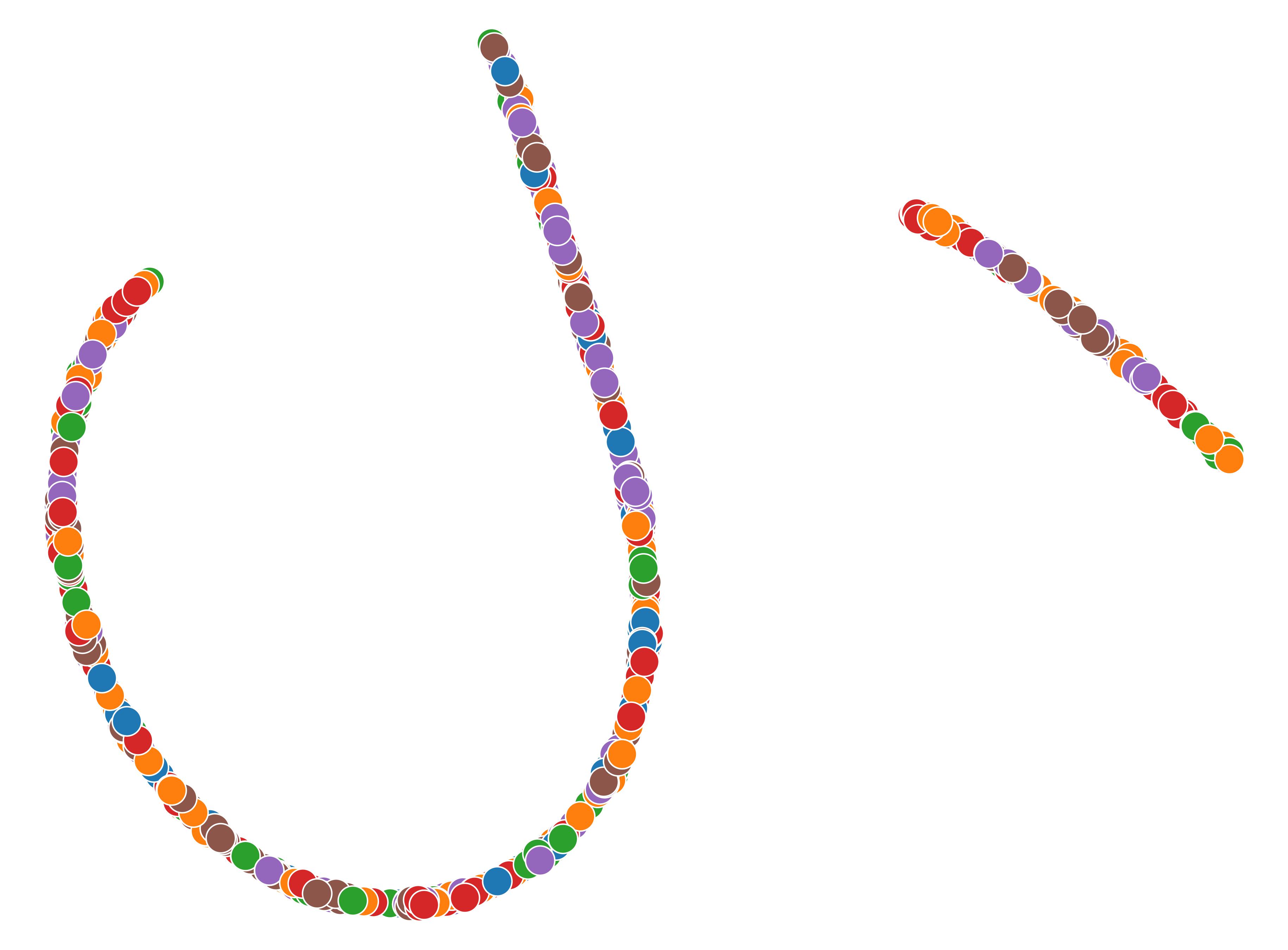}
                        &   \includegraphics[width=0.15\textwidth,valign=m]{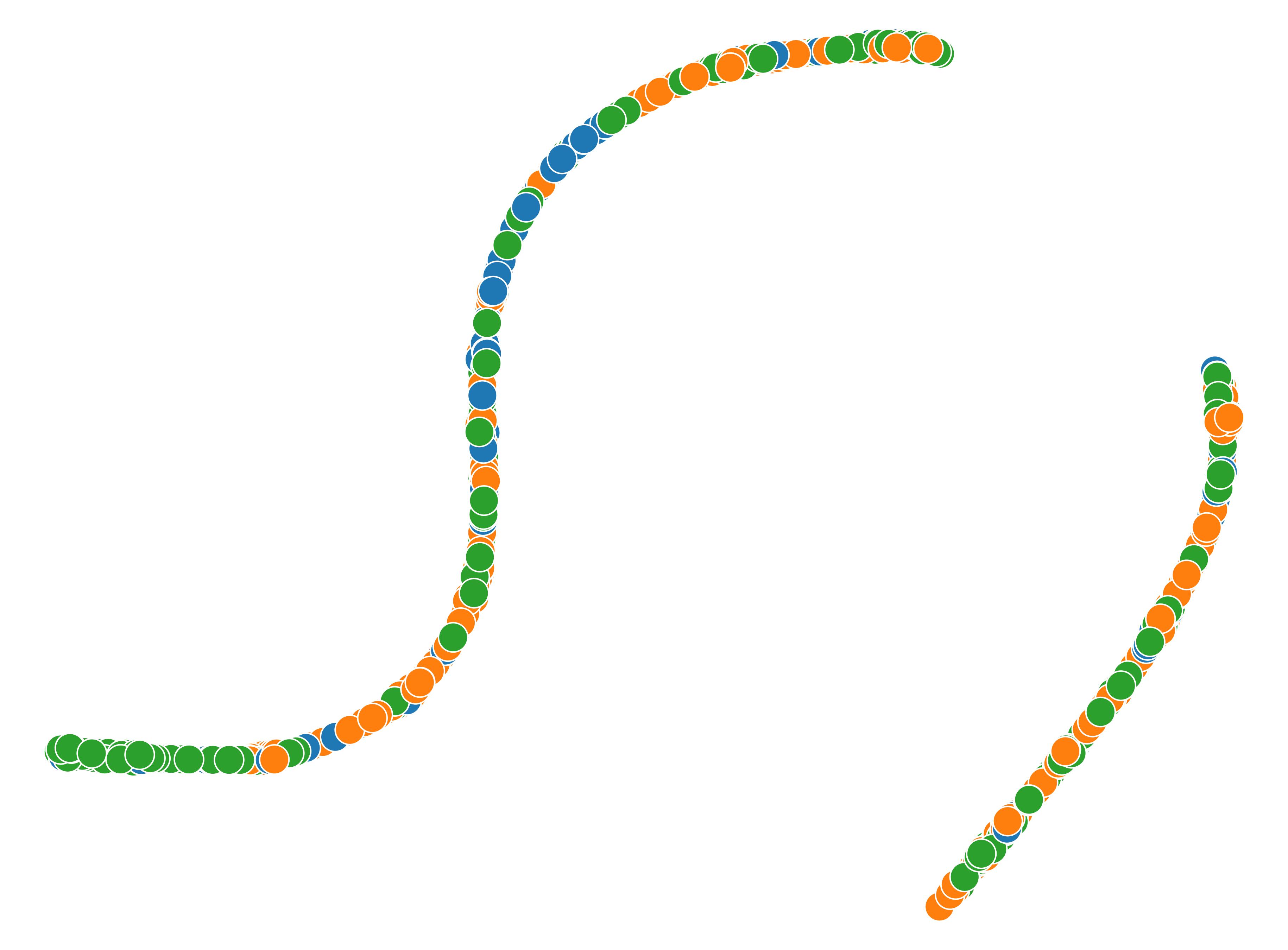}
                        \\
\midrule[0.1pt]
\rothead{\centering{UGTs-16}} 
                        &   \includegraphics[width=0.15\textwidth,valign=m]{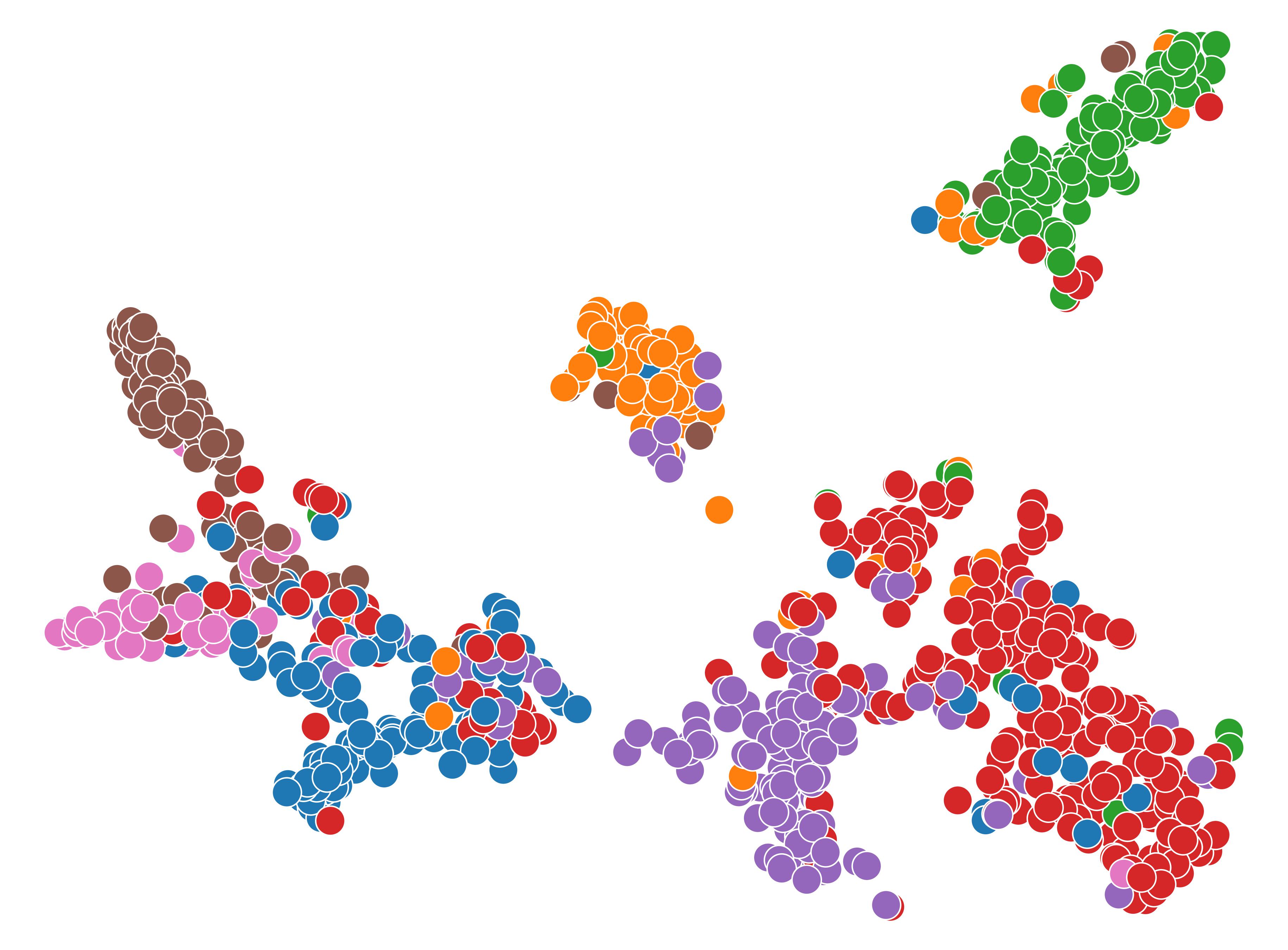}    
                        &   \includegraphics[width=0.15\textwidth,valign=m]{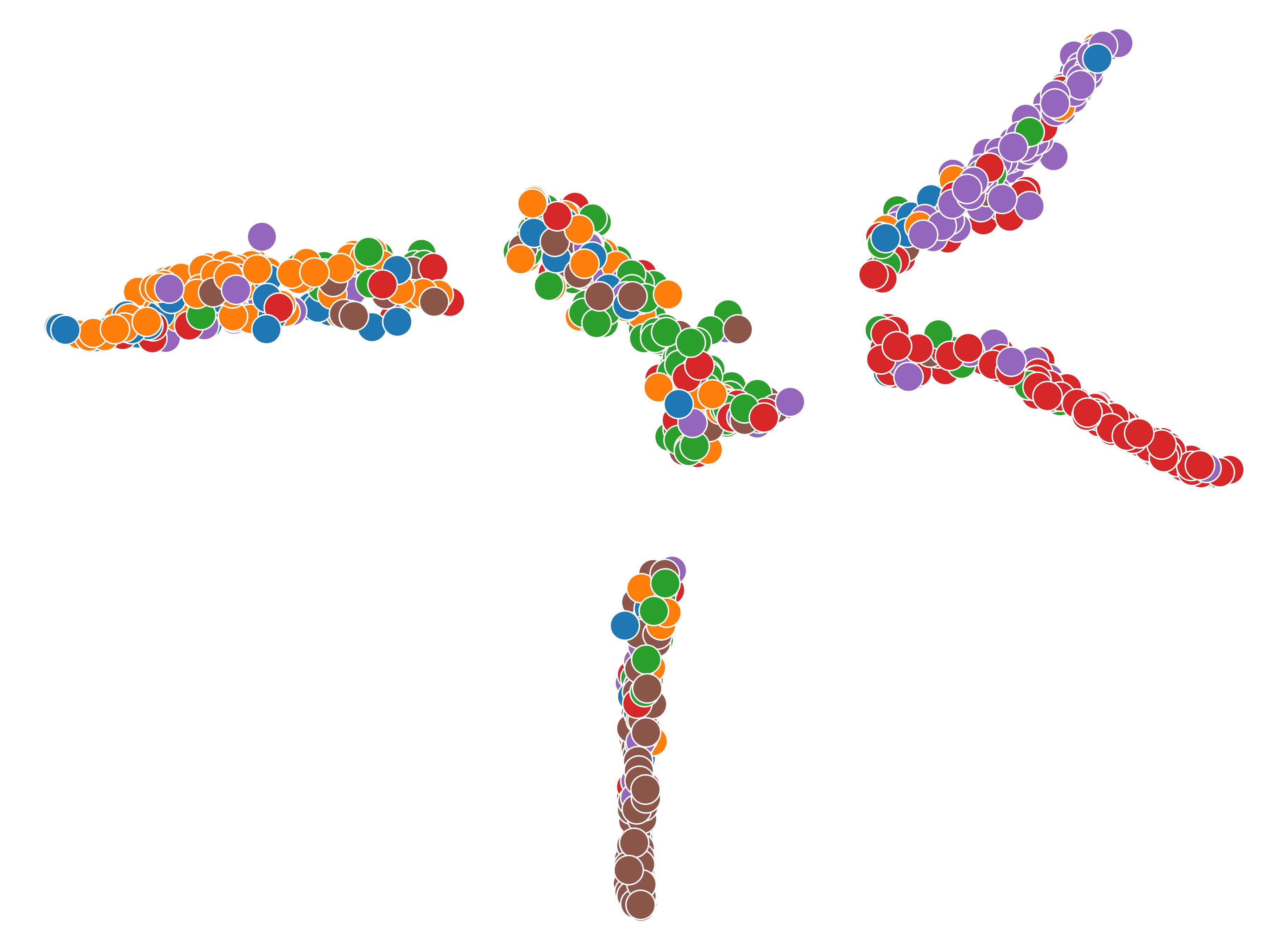}
                        &   \includegraphics[width=0.15\textwidth,valign=m]{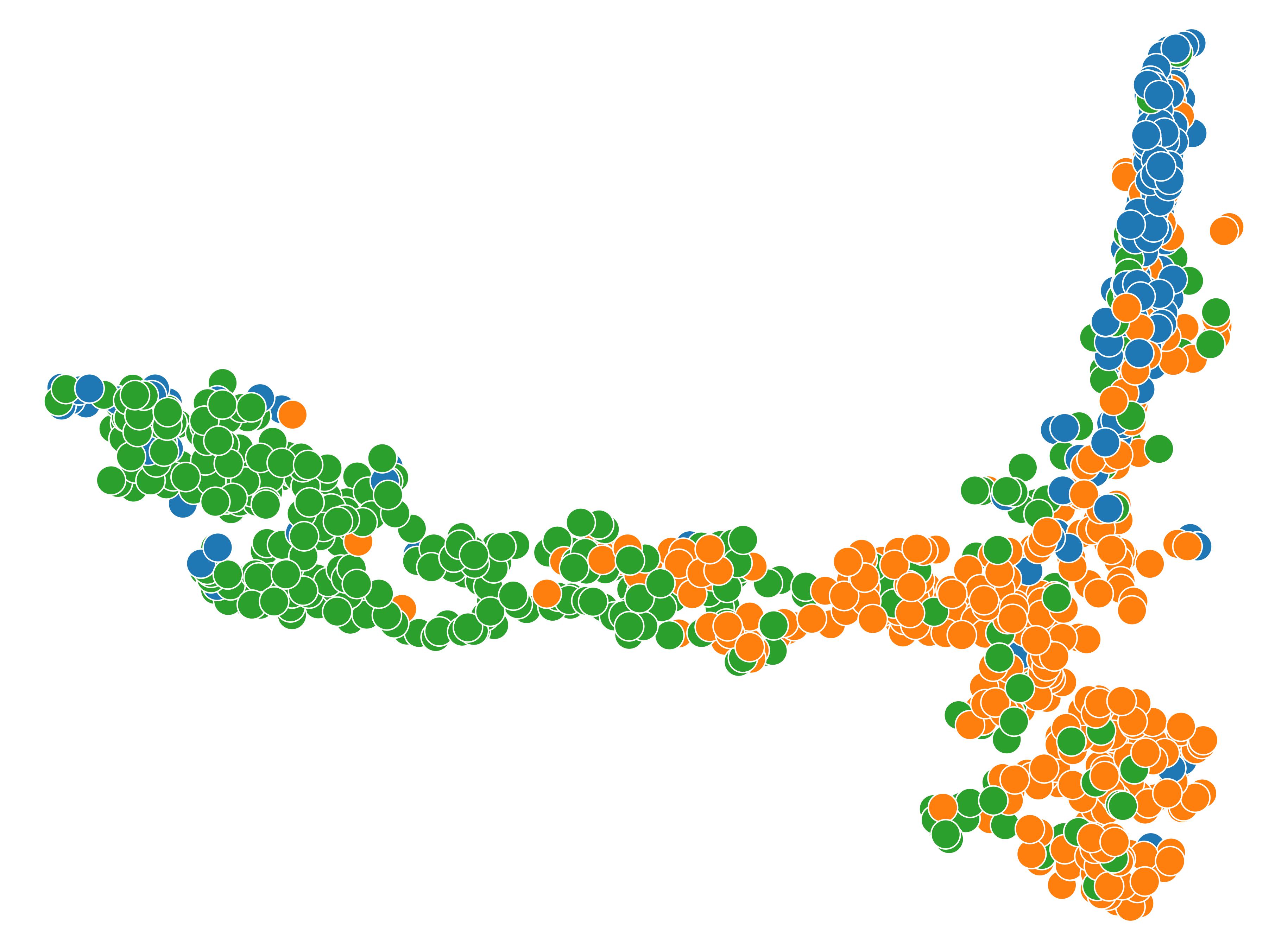} 
\rothead{\centering{UGTs-32}} 
                        &   \includegraphics[width=0.15\textwidth,valign=m]{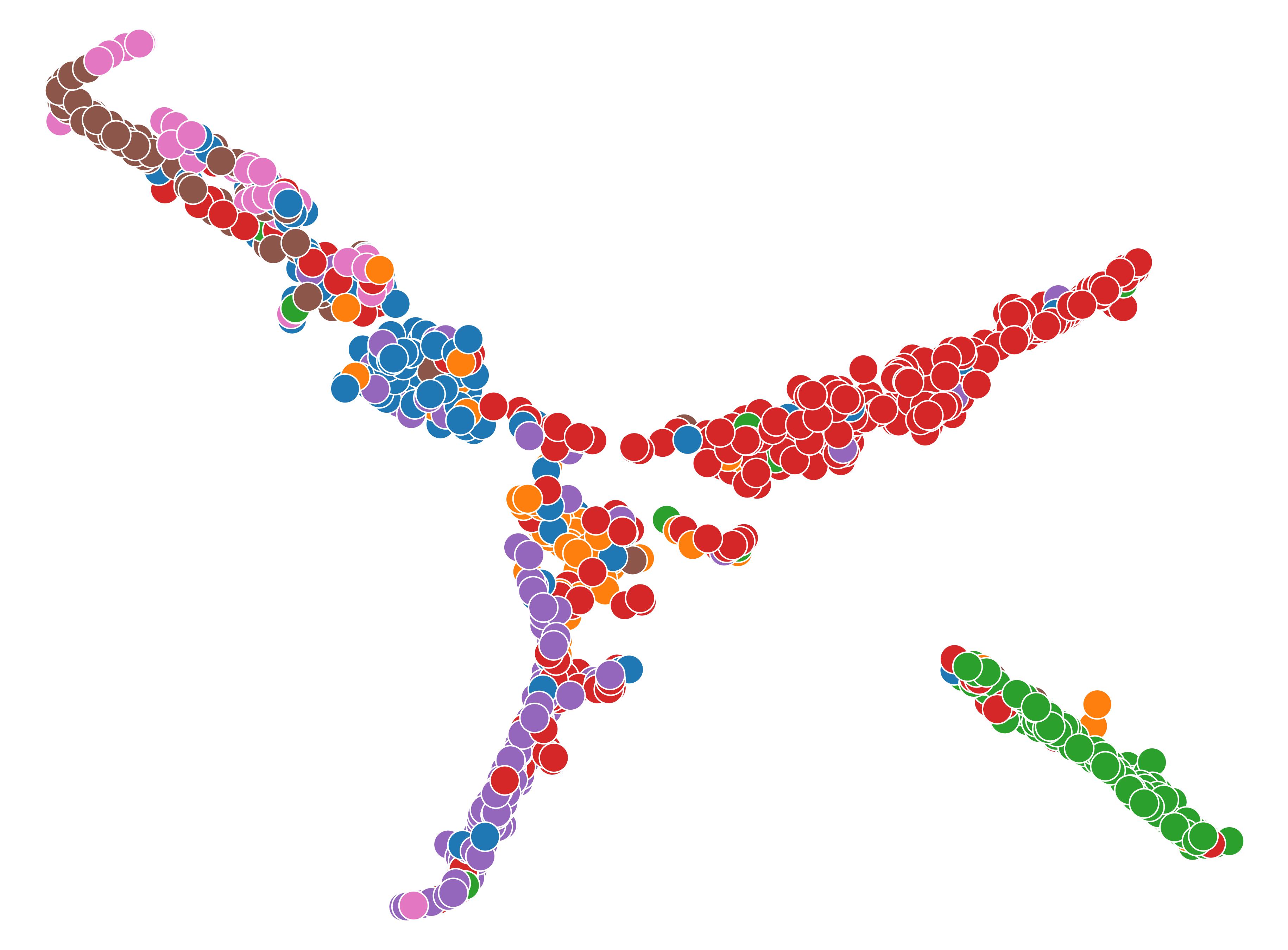}    
                        &   \includegraphics[width=0.15\textwidth,valign=m]{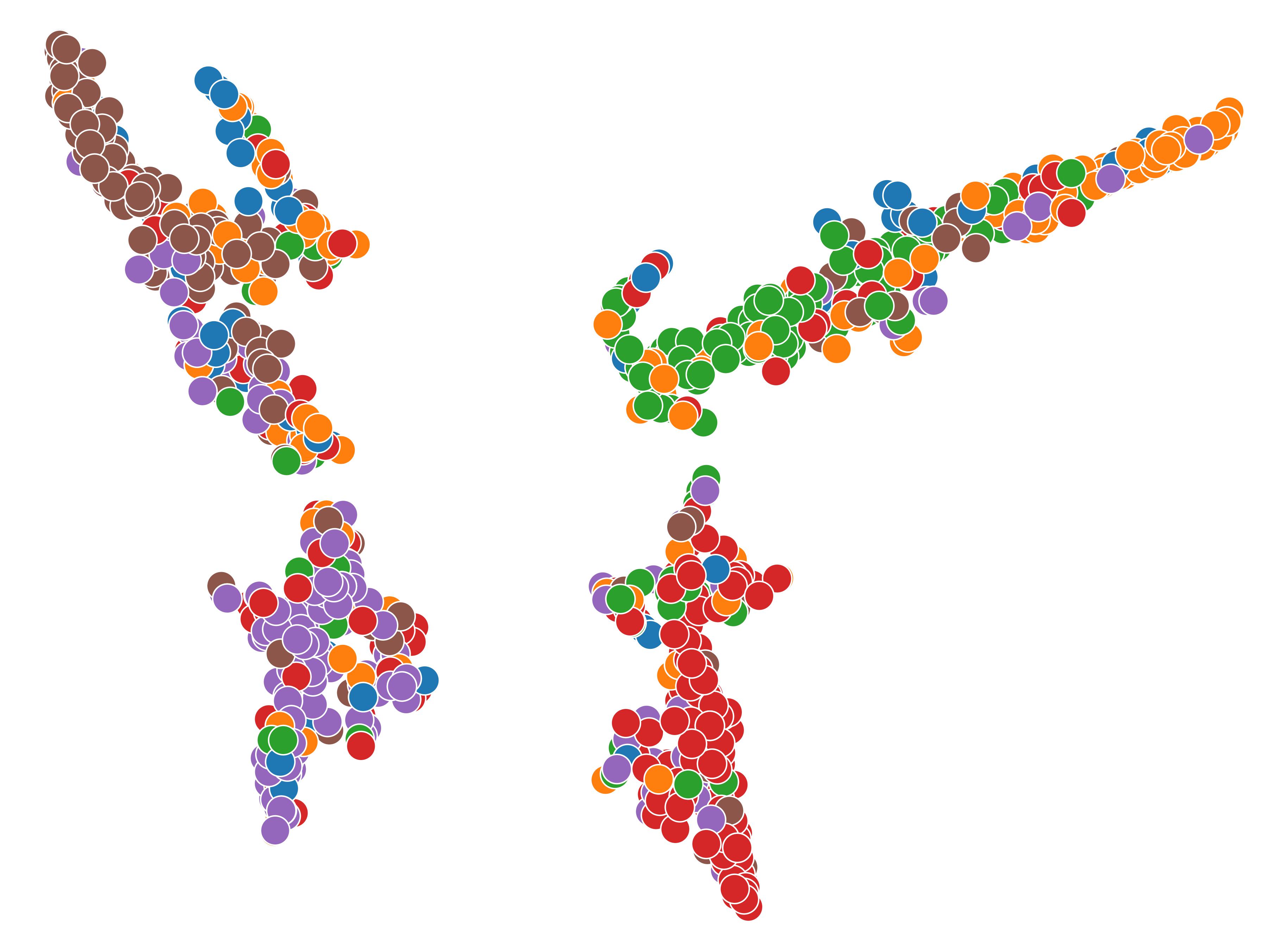}
                        &   \includegraphics[width=0.15\textwidth,valign=m]{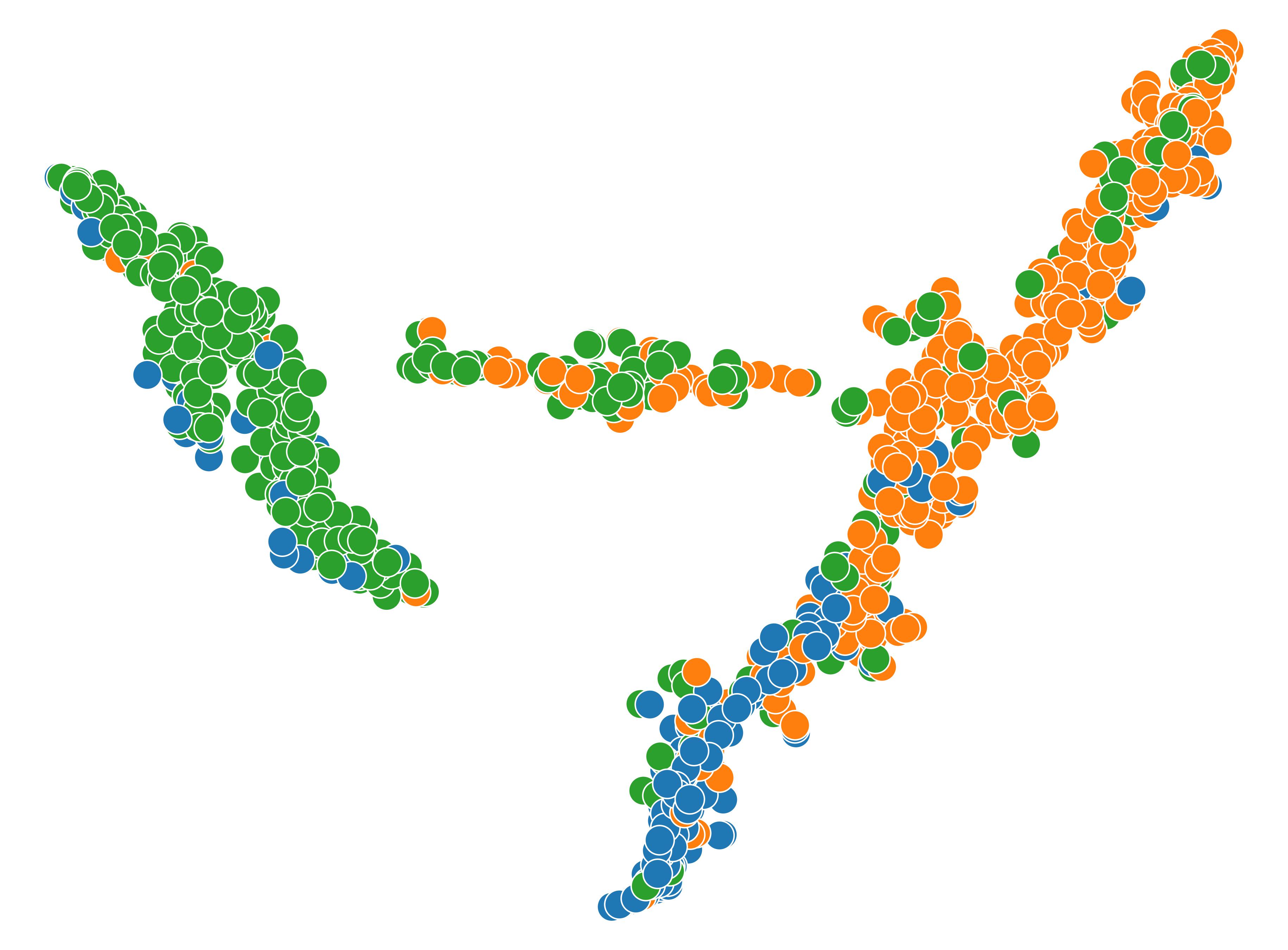}
                        \\
\midrule[2pt]
\end{tabularx}
\end{adjustbox}
\caption{TSNE visualization of node representations learned by densely trained GCN and UGTs. Ten classes are randomly sampled from OGBN-Arxiv for visualization. Model depth is set as 16 and 32 respectively; width is set as 448. See Appendix~\ref{appendix_tsne} for GAT architecture. }
\label{Fig:TSNE_GCN}
\end{figure*}

\subsection{Over-smoothing Analysis} 
Deep architecture has been shown as a key factor that improves the model capability in computer vision~\citep{he2016deep}. However, it becomes less appealing in GNNs mainly because the node interaction through the message-passing mechanism (i.e., aggregation operator) would make node representations less distinguishable~\citep{chen2020measuring,zhao2019pairnorm}, leading to a drastic drop of task performance. This phenomenon is well known as the over-smoothing problem~\citep{li2018deeper,chen2021bag}. In this paper, we show a surprising result that UGTs can effectively mitigate over-smoothing in deep GNNs. We conduct extensive experiments to evaluate this claim in this section.

\textbf{UGTs preserves the high accuracy as GNNs go deeper.} In Figure~\ref{fig:layers}, we vary the model depth of various architectures and report the test accuracy. All the experiments are conducted with architectures containing width 448 except for GAT on OGBN-Arxiv, in which we choose width 256 for GAT with $2\sim10$ layers and width 128 for GAT with $11\sim20$ layers, due to the memory limitation. 

As we can see, the performance of trained dense GNNs suffers from a sharp performance drop when the model goes deeper, whereas UGTs impressively preserves the high accuracy across models. Especially at the mild sparsity, i.e., 0.1, UGTs almost has no deterioration with the increased number of layers. 

\textbf{UGTs achieves competitive performance with the well-versed training techniques.}
To further validate the effectiveness of UGTs in mitigating over-smoothing, we compare UGTs with six  state-of-the-art techniques for the over-smoothing problem, including Residual connections, Jumping connections, NodeNorm, PairNorm, DropEdge, and DropNode. We follow the experimental setting in~\citep{chen2021bag} and conduct experiments on Cora/Citeseer/Pubmed with GAT containing 16 and 32 layers. Model width is set to 448 for GAT on Cora/Citeseer/Pubmed.  The results of the other methods are obtained from~\cite{chen2021bag}\footnote{\url{https://github.com/VITA-Group/Deep_GCN_Benchmarking.git}}. 

Table~\ref{tab:effective_oversmooth} shows that UGTs consistently outperforms all these advanced techniques on Cora, Citeseer, and Pubmed. For instance, UGTs outperforms the best performing technique (+Jumping) by 2.0\%, 1.2\%, 1.0\% on Cora, Citeseer and Pubmed respectively with 32 layers. These results again verify our hypothesis that training bottlenecks of deep GNNs (e.g., over-smoothing) can be avoided or mitigated by finding untrained subnetworks without training weights at all. 



\begin{figure}[h]
\centering
\subfloat{\includegraphics[width=0.99\textwidth]{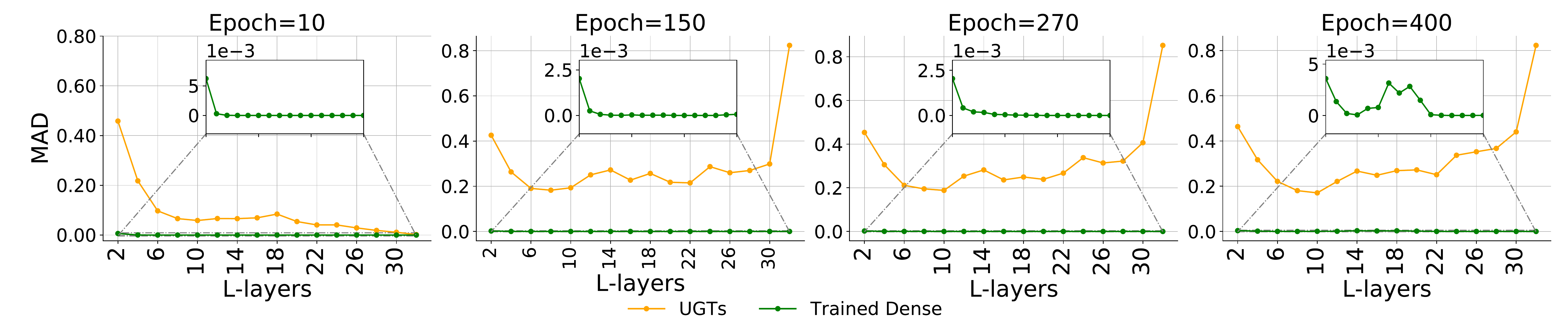} 
}
\caption{Mean Average Distance among node representations of each GNN layer. Experiments are conducted on Cora with GCN containing 32 layers and width 448.}
\label{fig:featureConsinDis}
\end{figure}

\begin{figure*}[!htb]
\centering
\subfloat{\includegraphics[width=0.99\textwidth]{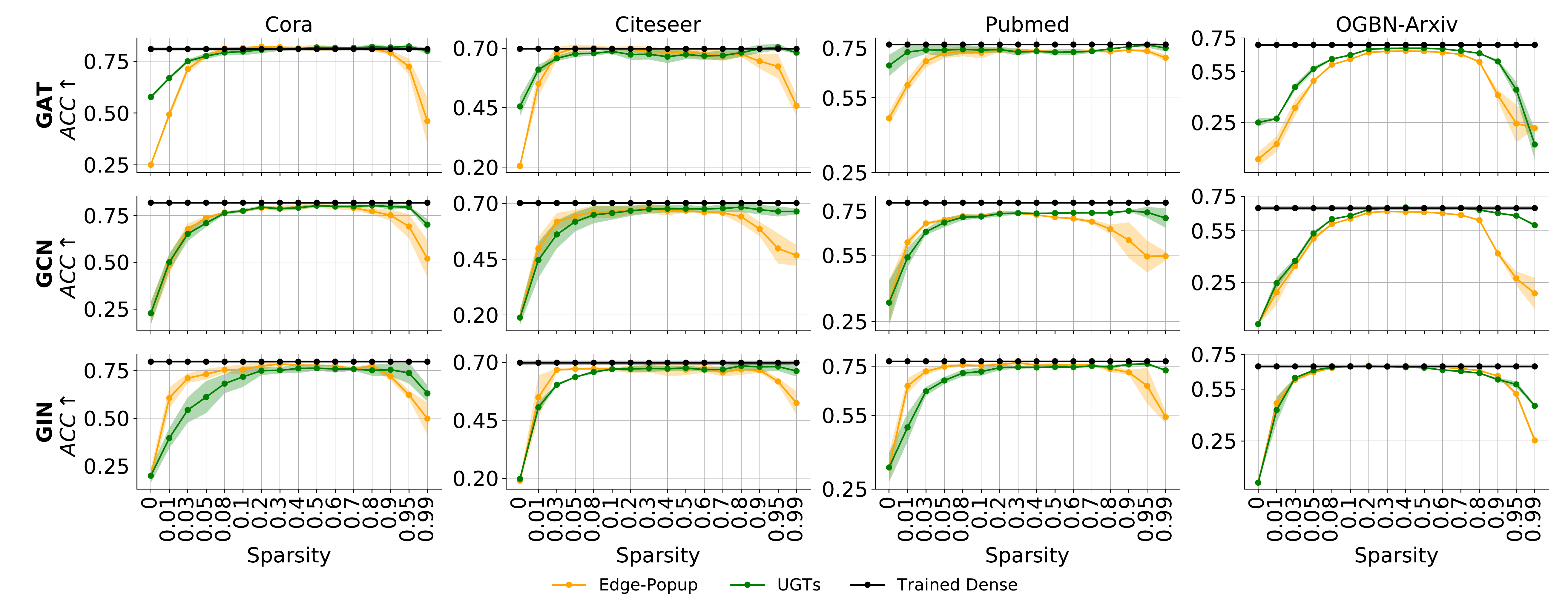}}
\caption{The accuracy of GNNs w.r.t varying sparsities. Experiments are conducted on various GNNs with 2 layers and width 256 for Cora, Citeseer and Pubmed, 4 layers and width 386 for OGBN-Arxiv. }
\label{fig:Sparsity}
\end{figure*}


\textbf{Mean Average Distance (MAD).} 
To further evaluate whether or not the good performance of UGTs can be contributed to the mitigation of over-smoothing, we visualize the smoothness of the node representations learned by UGTs and trained dense GNNs respectively. Following~\cite{chen2020measuring}, we calculate the MAD distance among node representations for each layer during the process of sparsification. Concretely, MAD~\citep{chen2020measuring} is the quantitative metric for measuring the smoothness of the node representations. The smaller the MAD is, the smoother the node representations are. Results are reported in Figure~\ref{fig:featureConsinDis}. It can be observed that the node representations learned by UGTs keeps having a large distance throughout the optimization process, indicating a relieving of over-smoothing.  On the contrary, the densely trained GCN suffers from severely indistinguishable representations of nodes.  

\textbf{TSNE Visualizations.} Additionally, we visualize the node representations learned by UGTs and the trained dense GNNs with 16 and 32 layers, respectively, on both GCN and GAT architectures. Due to the limited space, we show the results of GCN in Figure~\ref{Fig:TSNE_GCN} and put the visualization of GAT in the Appendix~\ref{appendix_tsne}. We can see that the node representations learned by the trained dense GCN are over-mixing in all scenarios and, in the deeper models (i.e., 32 layers), seem to be more indistinguishable. Meanwhile, the projection of node representations learned by UGTs maintains clearly distinguishable, again providing the empirical evidence of UGTs in mitigating over-smoothing problem.

\subsection{The Effect of Sparsity on UGTs}

To better understand the effect of sparsity on the performance of UGTs, we provide a comprehensive study in Figure~\ref{fig:Sparsity} where the performance of UGTs with respect to different sparsity levels on different architectures. We summarize our observations below. 

\textbf{\circled{1} UGTs consistently finds matching untrained graph subnetworks at a large range of sparsities, including the extreme ones.} A matching untrained graph subnetwork can be identified with sparsities from 0.1 even up to 0.99 on small-scale datasets such as Cora, Citeseer and Pubmed. For large-scale OGBN-Arxiv, it is more difficult to find matching untrained subnetworks. Matching subnetworks are mainly located within sparsities of 0.3 $\sim$ 0.6.

\textbf{\circled{2} What's more, UGTs consistently outperforms Edge-Popup.} UGTs shows better performance than Edge-Popup at high sparsities across different architectures on Cora, Citeseer, Pubmed and OGBN-Arxiv. Surprisingly, increasing sparsity from 0.7 to 0.99, UGTs maintains very a high accuracy, whereas the accuracy of Edge-Popup shows a notable degradation. It is in accord with our expectation since UGTs finds important weights globally by searching for the well-performing sparse topology across layers.   

\subsection{Broader Evaluation of UGTs}

\begin{figure*}[!htb]
\centering
\subfloat{\includegraphics[width=0.98\textwidth]{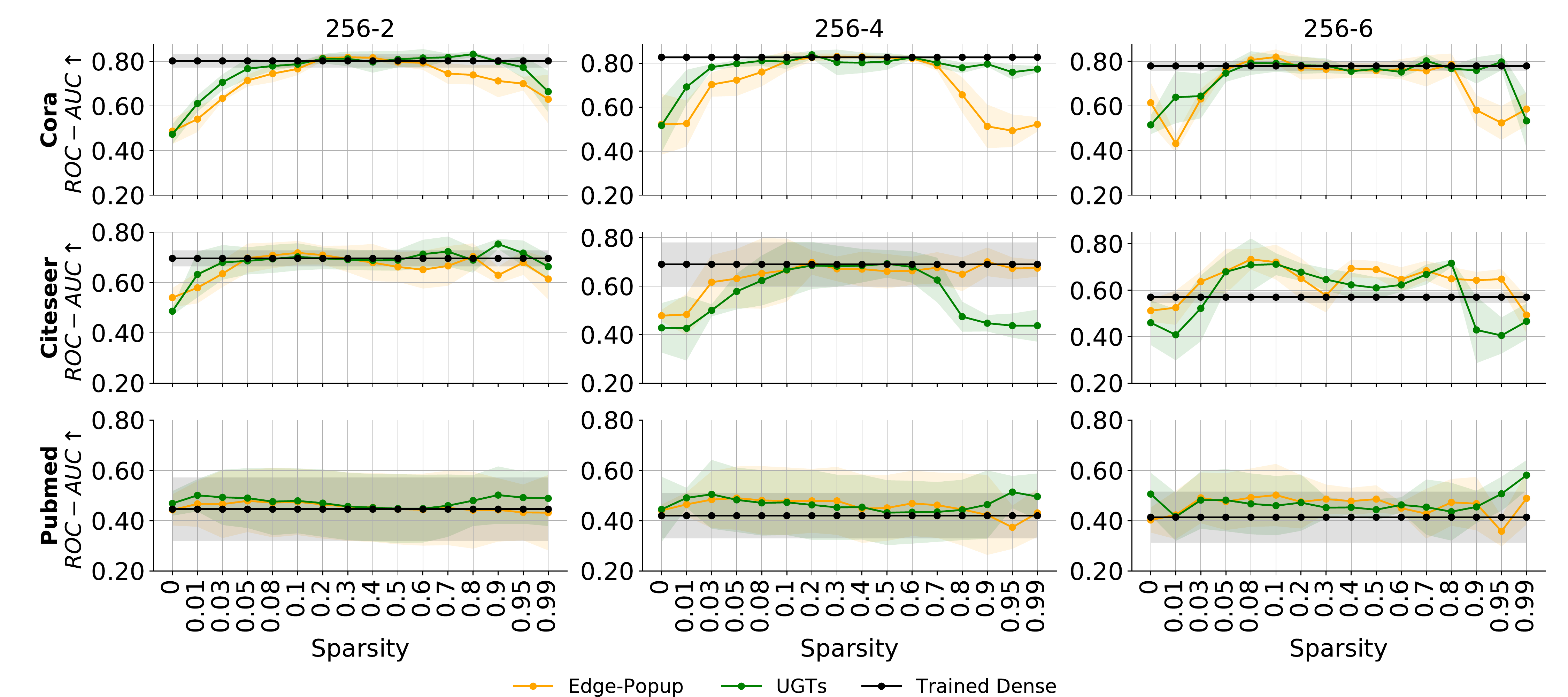}}
\caption{Out-of-distribution performance (ROC-AUC).  Experiments are conducted with GCN (Width: 256, Depth: 2).}
\label{fig:OOD_GCN}
\end{figure*}

\begin{figure*}[htb!]
\centering
\subfloat{\includegraphics[width=0.98\textwidth]{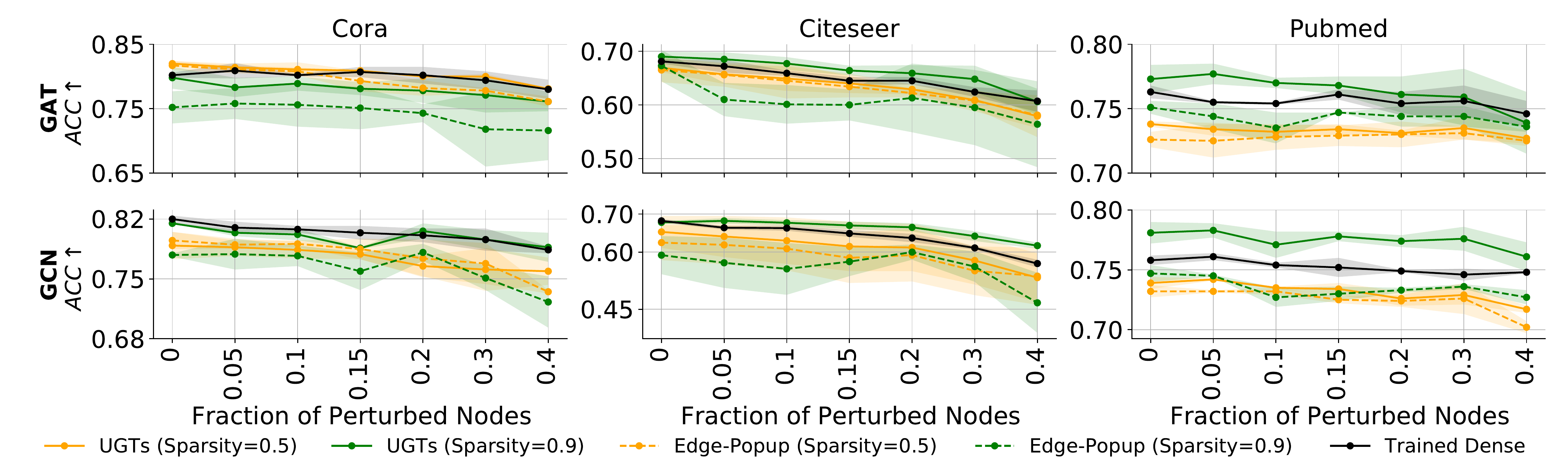}}
\caption{The robust performance on feature perturbations with the fraction of perturbed nodes varying from 0\% to 40\%. Experiments are conducted with GCN and GAT (Width: 256, Depth: 2).}
\label{fig:Robust_features_budgets}
\end{figure*}

\begin{figure*}[htb!]
\vspace{-1em}
\centering
\subfloat{\includegraphics[width=0.98\textwidth]{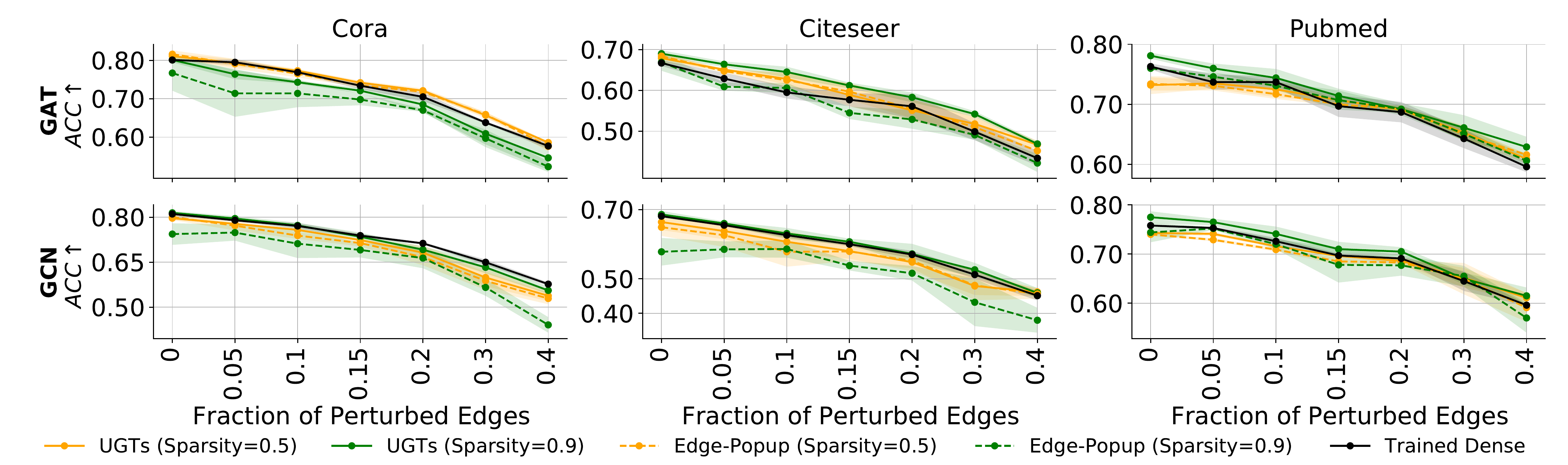}}
\caption{The robust performance on edge perturbations with the fraction of perturbed edges varying from 0\% to 40\%. Experiments are conducted with GCN and GAT (Width: 256, Depth: 2).}
\label{fig:Robust_Edges_budgets}
\vskip -0.1 in
\end{figure*}

In this section, we systematically study the performance of UGTs on out of distribution (OOD) detection, robustness against the input perturbations including feature and edge perturbations. Following ~\cite{stadler2021graph}, we create OOD samples by specifying all samples from 40\% of classes and removing them from the training set. We create feature perturbations by replacing them with the noise sampled from Bernoulli distribution with p=0.5 and edge perturbations by moving edge's end point at random. The results of OOD experiments are reported in Figure~\ref{fig:OOD_GCN} and Figure~\ref{fig:OOD_GAT} (shown in Appendix~\ref{appendix_OOD}). The results of robustness experiments are reported in Figure~\ref{fig:Robust_Edges_budgets} and Figure~\ref{fig:Robust_features_budgets}. We summarize our observations as follows:

\textbf{\circled{1} UGTs enjoys matching performance on OOD detection.} Figure~\ref{fig:OOD_GCN} and Figure~\ref{fig:OOD_GAT} show that untrained graph subnetworks discovered by UGTs achieve matching performance on OOD detection compared with the trained dense GNNs in most cases. Besides, UGTs consistently outperforms Edge-Popup method at a large range of sparsities on OOD detection. 

\textbf{\circled{2} UGTs produces highly sparse yet robust subnetworks on input perturbations.} Figure~\ref{fig:Robust_features_budgets} and Figure~\ref{fig:Robust_Edges_budgets} demonstrate that UGTs with high sparsity level (Sparsity=0.9) achieves more robust results than the trained dense GNNs on both feature and edge perturbations with perturbation percentage ranging from 0 to  40\%.  Again, UGTs consistently outperforms Edge-Popup with both perturbation types.

\begin{table*}[!htb]
\centering
\caption{Experments on graph-level tasks and other datasets. GCN Model with width:448, depth:3 are adopted for this experiments.}\label{tab:graph_otherdatasets}
\begin{adjustbox}{width=0.8\textwidth}
\begin{tabular}{c|c|c|c|c|c|c|c|c|c|c|c}
\toprule[2pt]
\multicolumn{12}{c}{Node-level: OGBN-Products (Accuracy:\%)}\\
\midrule[1.5pt]
Sparsity=&0.1&0.2&0.3&0.4&0.5&0.6&0.7&0.8&0.9&0.95&0.99\\
\midrule[2pt]
Dense&79.5&79.5&79.5&79.5&79.5&79.5&79.5&79.5&79.5&79.5&79.5\\
\textbf{UGTs}&75.6&77.7&78.7&77.8&79.3&79.5&79.9&78.6&74.5&64.1&35.3\\
\midrule[1pt]
\midrule[1pt]
\multicolumn{12}{c}{Node-level: TEXAS (Accuracy:\%)} \\
\midrule[1pt]
Dense&62.2&62.2&62.2&62.2&62.2&62.2&62.2&62.2&62.2&62.2&62.2\\
\textbf{UGTs}&62.1&62.2&62.2&62.2&62.2&61.3&62.2&63.1&64.8&64.8&55.8\\
\midrule[2pt]
\multicolumn{12}{c}{Graph-level: OGBG-molhiv (ROCAUC:\%)} \\
\midrule[1pt]
Dense&77.4&77.4&77.4&77.4&77.4&77.4&77.4&77.4&77.4&77.4&77.4\\
\textbf{UGTs}&76.4&76.9&76.5&76.1&76.3&77.3&75.8&77.1&73.1&75.3&75.1\\
\midrule[1pt]
\midrule[1pt]
\multicolumn{12}{c}{Graph-level: OGBG-molbace (ROCAUC:\%)} \\
\midrule[1pt]
Dense&78.3&78.3&78.3&78.3&78.3&78.3&78.3&78.3&78.3&78.3&78.3\\
\textbf{UGTs}&76.0&73.7&77.0&77.1&77.0&77.6&78.4&77.3&76.2&75.6&74.9\\
\bottomrule[2pt]
\end{tabular}
\end{adjustbox}
\vskip -0.1 in
\end{table*}

\subsection{Experiments on Graph-level Task and Other Datasets}
To draw a solid conclusion, we further conduct extensive experiments of graph-level task on  OGBG-molhiv and OGBG-molbace; node-level task on TEXAS and OGBN-Products. The experiments are based on GCN model with width=448 and depth=3. Table~\ref{tab:graph_otherdatasets} consistently verifies that a matching untrained subnetwork can be identified in GNNs across multiple tasks and datasets.

\section{Conclusion}
In this work, we for the first time confirm the existence of matching untrained subnetworks at a large range of sparsity. 
UGTs consistently outperforms the previous untrained technique -- Edge-Popup on multiple graph datasets across various GNN architectures. What's more, we show a surprising result that searching for an untrained subnetwork within a randomly weighted dense GNN instead of directly training the latter can significantly mitigate the over-smoothing problem of deep GNNs.  Across popular datasets, e.g., Cora, Citeseer, Pubmed, and OGBN-Arxiv, our method UGTs can achieve comparable or better performance with the various well-studied techniques that are specifically designed for over-smoothing. Moreover, we empirically find that UGTs also achieves appealing performance on other desirable aspects, such as out-of-distribution detection and robustness. The strong results of our paper point out a surprising but perhaps worth-a-try direction to obtain high-performing GNNs, i.e., finding the Untrained Tickets located within a randomly weighted dense GNN instead of training it.

\section*{Acknowledgements}
This work used the Dutch national e-infrastructure with the support of the SURF Cooperative using grant no. NWO-2021.060 and EINF-3214/L1.

\bibliographystyle{unsrtnat}
\bibliography{reference}

\begin{thebibliography}{48}
\providecommand{\natexlab}[1]{#1}
\providecommand{\url}[1]{\texttt{#1}}
\expandafter\ifx\csname urlstyle\endcsname\relax
  \providecommand{\doi}[1]{doi: #1}\else
  \providecommand{\doi}{doi: \begingroup \urlstyle{rm}\Url}\fi

\bibitem[Scarselli et~al.(2008)Scarselli, Gori, Tsoi, Hagenbuchner, and
  Monfardini]{scarselli2008graph}
Franco Scarselli, Marco Gori, Ah~Chung Tsoi, Markus Hagenbuchner, and Gabriele
  Monfardini.
\newblock The graph neural network model.
\newblock \emph{IEEE transactions on neural networks}, 20\penalty0
  (1):\penalty0 61--80, 2008.

\bibitem[Li et~al.(2016)Li, Tarlow, Brockschmidt, and Zemel]{li2015gated}
Yujia Li, Daniel Tarlow, Marc Brockschmidt, and Richard Zemel.
\newblock Gated graph sequence neural networks.
\newblock In \emph{International Conference on Learning Representations}, 2016.

\bibitem[Kipf and Welling(2017)]{kipf2016semi}
Thomas~N Kipf and Max Welling.
\newblock Semi-supervised classification with graph convolutional networks.
\newblock \emph{International Conference on Learning Representations}, 2017.

\bibitem[Xu et~al.(2019)Xu, Hu, Leskovec, and Jegelka]{xu2018how}
Keyulu Xu, Weihua Hu, Jure Leskovec, and Stefanie Jegelka.
\newblock How powerful are graph neural networks?
\newblock In \emph{International Conference on Learning Representations}, 2019.
\newblock URL \url{https://openreview.net/forum?id=ryGs6iA5Km}.

\bibitem[Veli{\v{c}}kovi{\'c} et~al.(2017)Veli{\v{c}}kovi{\'c}, Cucurull,
  Casanova, Romero, Lio, and Bengio]{velivckovic2017graph}
Petar Veli{\v{c}}kovi{\'c}, Guillem Cucurull, Arantxa Casanova, Adriana Romero,
  Pietro Lio, and Yoshua Bengio.
\newblock Graph attention networks.
\newblock \emph{arXiv preprint arXiv:1710.10903}, 2017.

\bibitem[Qiu et~al.(2018)Qiu, Tang, Ma, Dong, Wang, and Tang]{qiu2018deepinf}
Jiezhong Qiu, Jian Tang, Hao Ma, Yuxiao Dong, Kuansan Wang, and Jie Tang.
\newblock Deepinf: Social influence prediction with deep learning.
\newblock In \emph{Proceedings of the 24th ACM SIGKDD International Conference
  on Knowledge Discovery \& Data Mining}, pages 2110--2119, 2018.

\bibitem[Li et~al.(2019{\natexlab{a}})Li, Han, Cheng, Su, Wang, Zhang, and
  Pan]{li2019predicting}
Jia Li, Zhichao Han, Hong Cheng, Jiao Su, Pengyun Wang, Jianfeng Zhang, and
  Lujia Pan.
\newblock Predicting path failure in time-evolving graphs.
\newblock In \emph{Proceedings of the 25th ACM SIGKDD International Conference
  on Knowledge Discovery \& Data Mining}, pages 1279--1289, 2019{\natexlab{a}}.

\bibitem[Zitnik and Leskovec(2017)]{zitnik2017predicting}
Marinka Zitnik and Jure Leskovec.
\newblock Predicting multicellular function through multi-layer tissue
  networks.
\newblock \emph{Bioinformatics}, 33\penalty0 (14):\penalty0 i190--i198, 2017.

\bibitem[Guo et~al.(2019)Guo, Lin, Feng, Song, and Wan]{guo2019attention}
Shengnan Guo, Youfang Lin, Ning Feng, Chao Song, and Huaiyu Wan.
\newblock Attention based spatial-temporal graph convolutional networks for
  traffic flow forecasting.
\newblock In \emph{Proceedings of the AAAI Conference on Artificial
  Intelligence}, volume~33, pages 922--929, 2019.

\bibitem[Ying et~al.(2018)Ying, He, Chen, Eksombatchai, Hamilton, and
  Leskovec]{ying2018graph}
Rex Ying, Ruining He, Kaifeng Chen, Pong Eksombatchai, William~L Hamilton, and
  Jure Leskovec.
\newblock Graph convolutional neural networks for web-scale recommender
  systems.
\newblock In \emph{Proceedings of the 24th ACM SIGKDD International Conference
  on Knowledge Discovery \& Data Mining}, pages 974--983, 2018.

\bibitem[Zhou et~al.(2019)Zhou, Lan, Liu, and Yosinski]{zhou2019deconstructing}
Hattie Zhou, Janice Lan, Rosanne Liu, and Jason Yosinski.
\newblock Deconstructing lottery tickets: Zeros, signs, and the supermask.
\newblock \emph{arXiv preprint arXiv:1905.01067}, 2019.

\bibitem[Ramanujan et~al.(2020)Ramanujan, Wortsman, Kembhavi, Farhadi, and
  Rastegari]{ramanujan2020s}
Vivek Ramanujan, Mitchell Wortsman, Aniruddha Kembhavi, Ali Farhadi, and
  Mohammad Rastegari.
\newblock What's hidden in a randomly weighted neural network?
\newblock In \emph{Proceedings of the IEEE/CVF Conference on Computer Vision
  and Pattern Recognition}, pages 11893--11902, 2020.

\bibitem[Chijiwa et~al.(2021)Chijiwa, Yamaguchi, Ida, Umakoshi, and
  Inoue]{chijiwa2021pruning}
Daiki Chijiwa, Shin'ya Yamaguchi, Yasutoshi Ida, Kenji Umakoshi, and Tomohiro
  Inoue.
\newblock Pruning randomly initialized neural networks with iterative
  randomization.
\newblock \emph{arXiv preprint arXiv:2106.09269}, 2021.

\bibitem[Li et~al.(2018)Li, Han, and Wu]{li2018deeper}
Qimai Li, Zhichao Han, and Xiao-Ming Wu.
\newblock Deeper insights into graph convolutional networks for semi-supervised
  learning.
\newblock In \emph{Thirty-Second AAAI conference on artificial intelligence},
  2018.

\bibitem[Hu et~al.(2020)Hu, Fey, Zitnik, Dong, Ren, Liu, Catasta, and
  Leskovec]{hu2020open}
Weihua Hu, Matthias Fey, Marinka Zitnik, Yuxiao Dong, Hongyu Ren, Bowen Liu,
  Michele Catasta, and Jure Leskovec.
\newblock Open graph benchmark: Datasets for machine learning on graphs.
\newblock \emph{arXiv preprint arXiv:2005.00687}, 2020.

\bibitem[Frankle and Carbin(2018)]{frankle2018lottery}
Jonathan Frankle and Michael Carbin.
\newblock The lottery ticket hypothesis: Finding sparse, trainable neural
  networks.
\newblock \emph{arXiv preprint arXiv:1803.03635}, 2018.

\bibitem[Chen et~al.(2021{\natexlab{a}})Chen, Sui, Chen, Zhang, and
  Wang]{chen2021unified}
Tianlong Chen, Yongduo Sui, Xuxi Chen, Aston Zhang, and Zhangyang Wang.
\newblock A unified lottery ticket hypothesis for graph neural networks.
\newblock In \emph{International Conference on Machine Learning}, pages
  1695--1706. PMLR, 2021{\natexlab{a}}.

\bibitem[Zhu and Gupta(2017)]{zhu2017prune}
Michael Zhu and Suyog Gupta.
\newblock To prune, or not to prune: exploring the efficacy of pruning for
  model compression.
\newblock \emph{arXiv preprint arXiv:1710.01878}, 2017.

\bibitem[Gale et~al.(2019)Gale, Elsen, and Hooker]{gale2019state}
Trevor Gale, Erich Elsen, and Sara Hooker.
\newblock The state of sparsity in deep neural networks.
\newblock \emph{arXiv preprint arXiv:1902.09574}, 2019.

\bibitem[Liu et~al.(2021)Liu, Chen, Chen, Atashgahi, Yin, Kou, Shen,
  Pechenizkiy, Wang, and Mocanu]{liu2021sparse}
Shiwei Liu, Tianlong Chen, Xiaohan Chen, Zahra Atashgahi, Lu~Yin, Huanyu Kou,
  Li~Shen, Mykola Pechenizkiy, Zhangyang Wang, and Decebal~Constantin Mocanu.
\newblock Sparse training via boosting pruning plasticity with
  neuroregeneration.
\newblock \emph{Advances in Neural Information Processing Systems.}, 2021.

\bibitem[Hamilton et~al.(2017{\natexlab{a}})Hamilton, Ying, and
  Leskovec]{hamilton2017inductive}
William~L Hamilton, Rex Ying, and Jure Leskovec.
\newblock Inductive representation learning on large graphs.
\newblock In \emph{Proceedings of the 31st International Conference on Neural
  Information Processing Systems}, pages 1025--1035, 2017{\natexlab{a}}.

\bibitem[Wu et~al.(2019)Wu, Souza, Zhang, Fifty, Yu, and
  Weinberger]{wu2019simplifying}
Felix Wu, Amauri Souza, Tianyi Zhang, Christopher Fifty, Tao Yu, and Kilian
  Weinberger.
\newblock Simplifying graph convolutional networks.
\newblock In \emph{International conference on machine learning}, pages
  6861--6871. PMLR, 2019.

\bibitem[Kipf and Welling(2016)]{kipf2016variational}
Thomas~N Kipf and Max Welling.
\newblock Variational graph auto-encoders.
\newblock \emph{arXiv preprint arXiv:1611.07308}, 2016.

\bibitem[Li et~al.(2019{\natexlab{b}})Li, Muller, Thabet, and
  Ghanem]{li2019deepgcns}
Guohao Li, Matthias Muller, Ali Thabet, and Bernard Ghanem.
\newblock Deepgcns: Can gcns go as deep as cnns?
\newblock In \emph{Proceedings of the IEEE/CVF International Conference on
  Computer Vision}, pages 9267--9276, 2019{\natexlab{b}}.

\bibitem[Gong et~al.(2020)Gong, Bahri, Bronstein, and
  Zafeiriou]{gong2020geometrically}
Shunwang Gong, Mehdi Bahri, Michael~M Bronstein, and Stefanos Zafeiriou.
\newblock Geometrically principled connections in graph neural networks.
\newblock In \emph{Proceedings of the IEEE/CVF Conference on Computer Vision
  and Pattern Recognition}, pages 11415--11424, 2020.

\bibitem[Zhao and Akoglu(2019)]{zhao2019pairnorm}
Lingxiao Zhao and Leman Akoglu.
\newblock Pairnorm: Tackling oversmoothing in gnns.
\newblock \emph{arXiv preprint arXiv:1909.12223}, 2019.

\bibitem[Xu et~al.(2018)Xu, Li, Tian, Sonobe, Kawarabayashi, and
  Jegelka]{xu2018representation}
Keyulu Xu, Chengtao Li, Yonglong Tian, Tomohiro Sonobe, Ken-ichi Kawarabayashi,
  and Stefanie Jegelka.
\newblock Representation learning on graphs with jumping knowledge networks.
\newblock In \emph{International Conference on Machine Learning}, pages
  5453--5462. PMLR, 2018.

\bibitem[Liu et~al.(2020{\natexlab{a}})Liu, Gao, and Ji]{liu2020towards}
Meng Liu, Hongyang Gao, and Shuiwang Ji.
\newblock Towards deeper graph neural networks.
\newblock In \emph{Proceedings of the 26th ACM SIGKDD International Conference
  on Knowledge Discovery \& Data Mining}, pages 338--348, 2020{\natexlab{a}}.

\bibitem[Chen et~al.(2020{\natexlab{a}})Chen, Wei, Huang, Ding, and
  Li]{chen2020simple}
Ming Chen, Zhewei Wei, Zengfeng Huang, Bolin Ding, and Yaliang Li.
\newblock Simple and deep graph convolutional networks.
\newblock In \emph{International Conference on Machine Learning}, pages
  1725--1735. PMLR, 2020{\natexlab{a}}.

\bibitem[Zhou et~al.(2020)Zhou, Dong, Wang, Lee, Hooi, Xu, and
  Feng]{zhou2020understanding}
Kuangqi Zhou, Yanfei Dong, Kaixin Wang, Wee~Sun Lee, Bryan Hooi, Huan Xu, and
  Jiashi Feng.
\newblock Understanding and resolving performance degradation in graph
  convolutional networks.
\newblock \emph{arXiv preprint arXiv:2006.07107}, 2020.

\bibitem[Huang et~al.(2020)Huang, Rong, Xu, Sun, and Huang]{huang2020tackling}
Wenbing Huang, Yu~Rong, Tingyang Xu, Fuchun Sun, and Junzhou Huang.
\newblock Tackling over-smoothing for general graph convolutional networks.
\newblock \emph{arXiv preprint arXiv:2008.09864}, 2020.

\bibitem[Rong et~al.(2019)Rong, Huang, Xu, and Huang]{rong2019dropedge}
Yu~Rong, Wenbing Huang, Tingyang Xu, and Junzhou Huang.
\newblock Dropedge: Towards deep graph convolutional networks on node
  classification.
\newblock \emph{arXiv preprint arXiv:1907.10903}, 2019.

\bibitem[Wortsman et~al.(2020)Wortsman, Ramanujan, Liu, Kembhavi, Rastegari,
  Yosinski, and Farhadi]{wortsman2020supermasks}
Mitchell Wortsman, Vivek Ramanujan, Rosanne Liu, Aniruddha Kembhavi, Mohammad
  Rastegari, Jason Yosinski, and Ali Farhadi.
\newblock Supermasks in superposition.
\newblock \emph{arXiv preprint arXiv:2006.14769}, 2020.

\bibitem[Fu et~al.(2021)Fu, Yu, Zhang, Wu, Ouyang, Cox, and Lin]{fu2021drawing}
Yonggan Fu, Qixuan Yu, Yang Zhang, Shang Wu, Xu~Ouyang, David Cox, and Yingyan
  Lin.
\newblock Drawing robust scratch tickets: Subnetworks with inborn robustness
  are found within randomly initialized networks.
\newblock \emph{Advances in Neural Information Processing Systems}, 34, 2021.

\bibitem[Hamilton et~al.(2017{\natexlab{b}})Hamilton, Ying, and
  Leskovec]{hamilton2017representation}
William~L Hamilton, Rex Ying, and Jure Leskovec.
\newblock Representation learning on graphs: Methods and applications.
\newblock \emph{arXiv preprint arXiv:1709.05584}, 2017{\natexlab{b}}.

\bibitem[Bengio et~al.(2013)Bengio, L{\'e}onard, and
  Courville]{bengio2013estimating}
Yoshua Bengio, Nicholas L{\'e}onard, and Aaron Courville.
\newblock Estimating or propagating gradients through stochastic neurons for
  conditional computation.
\newblock \emph{arXiv preprint arXiv:1308.3432}, 2013.

\bibitem[Mocanu et~al.(2018)Mocanu, Mocanu, Stone, Nguyen, Gibescu, and
  Liotta]{mocanu2018scalable}
Decebal~Constantin Mocanu, Elena Mocanu, Peter Stone, Phuong~H Nguyen,
  Madeleine Gibescu, and Antonio Liotta.
\newblock Scalable training of artificial neural networks with adaptive sparse
  connectivity inspired by network science.
\newblock \emph{Nature communications}, 9\penalty0 (1):\penalty0 1--12, 2018.

\bibitem[Liu et~al.(2020{\natexlab{b}})Liu, Mocanu, Matavalam, Pei, and
  Pechenizkiy]{onemillionneurons}
Shiwei Liu, Decebal~Constantin Mocanu, Amarsagar Reddy~Ramapuram Matavalam,
  Yulong Pei, and Mykola Pechenizkiy.
\newblock Sparse evolutionary deep learning with over one million artificial
  neurons on commodity hardware.
\newblock \emph{Neural Computing and Applications}, 2020{\natexlab{b}}.

\bibitem[Evci et~al.(2020)Evci, Gale, Menick, Castro, and
  Elsen]{evci2020rigging}
Utku Evci, Trevor Gale, Jacob Menick, Pablo~Samuel Castro, and Erich Elsen.
\newblock Rigging the lottery: Making all tickets winners.
\newblock In \emph{International Conference on Machine Learning}, pages
  2943--2952. PMLR, 2020.

\bibitem[Kusupati et~al.(2020)Kusupati, Ramanujan, Somani, Wortsman, Jain,
  Kakade, and Farhadi]{kusupati2020soft}
Aditya Kusupati, Vivek Ramanujan, Raghav Somani, Mitchell Wortsman, Prateek
  Jain, Sham Kakade, and Ali Farhadi.
\newblock Soft threshold weight reparameterization for learnable sparsity.
\newblock In \emph{International Conference on Machine Learning}, 2020.

\bibitem[Dettmers and Zettlemoyer(2019)]{dettmers2019sparse}
Tim Dettmers and Luke Zettlemoyer.
\newblock Sparse networks from scratch: Faster training without losing
  performance.
\newblock \emph{arXiv preprint arXiv:1907.04840}, 2019.

\bibitem[Chen et~al.(2021{\natexlab{b}})Chen, Zhou, Duan, Zheng, Wang, Hu, and
  Wang]{chen2021bag}
Tianlong Chen, Kaixiong Zhou, Keyu Duan, Wenqing Zheng, Peihao Wang, Xia Hu,
  and Zhangyang Wang.
\newblock Bag of tricks for training deeper graph neural networks: A
  comprehensive benchmark study.
\newblock \emph{arXiv preprint arXiv:2108.10521}, 2021{\natexlab{b}}.

\bibitem[Pei et~al.(2020)Pei, Wei, Chang, Lei, and Yang]{pei2020geom}
Hongbin Pei, Bingzhe Wei, Kevin Chen-Chuan Chang, Yu~Lei, and Bo~Yang.
\newblock Geom-gcn: Geometric graph convolutional networks.
\newblock \emph{arXiv preprint arXiv:2002.05287}, 2020.

\bibitem[Wu et~al.(2018)Wu, Ramsundar, Feinberg, Gomes, Geniesse, Pappu,
  Leswing, and Pande]{wu2018moleculenet}
Zhenqin Wu, Bharath Ramsundar, Evan~N Feinberg, Joseph Gomes, Caleb Geniesse,
  Aneesh~S Pappu, Karl Leswing, and Vijay Pande.
\newblock Moleculenet: a benchmark for molecular machine learning.
\newblock \emph{Chemical science}, 9\penalty0 (2):\penalty0 513--530, 2018.

\bibitem[He et~al.(2016)He, Zhang, Ren, and Sun]{he2016deep}
Kaiming He, Xiangyu Zhang, Shaoqing Ren, and Jian Sun.
\newblock Deep residual learning for image recognition.
\newblock In \emph{Proceedings of the IEEE conference on computer vision and
  pattern recognition}, pages 770--778, 2016.

\bibitem[Chen et~al.(2020{\natexlab{b}})Chen, Lin, Li, Li, Zhou, and
  Sun]{chen2020measuring}
Deli Chen, Yankai Lin, Wei Li, Peng Li, Jie Zhou, and Xu~Sun.
\newblock Measuring and relieving the over-smoothing problem for graph neural
  networks from the topological view.
\newblock In \emph{Proceedings of the AAAI Conference on Artificial
  Intelligence}, volume~34, pages 3438--3445, 2020{\natexlab{b}}.

\bibitem[Stadler et~al.(2021)Stadler, Charpentier, Geisler, Z{\"u}gner, and
  G{\"u}nnemann]{stadler2021graph}
Maximilian Stadler, Bertrand Charpentier, Simon Geisler, Daniel Z{\"u}gner, and
  Stephan G{\"u}nnemann.
\newblock Graph posterior network: Bayesian predictive uncertainty for node
  classification.
\newblock \emph{Advances in Neural Information Processing Systems}, 34, 2021.

\bibitem[Ying et~al.(2021)Ying, Cai, Luo, Zheng, Ke, He, Shen, and
  Liu]{ying2021transformers}
Chengxuan Ying, Tianle Cai, Shengjie Luo, Shuxin Zheng, Guolin Ke, Di~He,
  Yanming Shen, and Tie-Yan Liu.
\newblock Do transformers really perform bad for graph representation?
\newblock \emph{arXiv preprint arXiv:2106.05234}, 2021.

\end{thebibliography}

\appendix
\clearpage
\section{Implementation Details} \label{implementDetails}

In this paper, all experiments on Cora/Citeseer/Pubmed datasets are conducted on 1 GeForce RTX 2080TI (11GB) and all experiments on OGBN-Arxiv are conducted on 1 DGX-A100 (40GB). All the results reported in this paper are conducted by 5 independent repeated runs. 

\textbf{Train-Val-Test splitting Datasets} We use 140 (Cora), 120 (Citeseer) and 60 (PubMed) labeled data for training, 500 nodes for validation and 1000 nodes for testing. We follow the strategy in~\cite{hu2020open} for splitting OGBN-Arxiv dataset.

\textbf{Hyper-parameter Configuration} We follow~\cite{chen2021bag,kipf2016semi,ying2021transformers} to configure the hyper-parameters for training dense GNN models. All  hyper-parameters configurations for UGTs  are summarized in Table~\ref{tab:train_details}.
\begin{table}[!htb]
\newcommand{\tabincell}[2]{\begin{tabular}{@{}#1@{}}#2\end{tabular}}
\centering
\caption{Implementation details for UGTs.}\label{datasets1}
\begin{adjustbox}{width=0.5\textwidth}
\begin{tabular}{l|c|c|c|c}
\toprule[2pt]
DataSets & Cora &Citeseer &Pubmed &OGBN-Arxiv \\
\midrule[2pt]
Total Epoches &400   & 400 &400&400\\
Learning Rate&0.01 & 0.01&0.01 &\tabincell{c}{0.01 (GNNs with Layers$<$10)\\0.001(GNNs with Layers$>$10)}\\
Optimizer &Adam &Adam&Adam &Adam \\
Weight Decay & 0.0 &0.0&0.0&0.0\\
n(total adjustion epoches) &200& 200&200&200\\
$t_{0}$ &0& 0& 0&0\\
$\Delta t$&1 epoch&1 epoch&1 epoch&1 epoch\\
\bottomrule[2pt]
\end{tabular}
\end{adjustbox}
\vspace{-0.1in}
\label{tab:train_details}
\end{table}

\section{More Experimental Results}

\subsection{TSNE visualization.}\label{appendix_tsne}
\begin{figure*}[!h]
\centering
\setlength\tabcolsep{1pt}
\settowidth\rotheadsize{Radcliffe Cam}
\begin{adjustbox}{width=0.9\textwidth}
\begin{tabularx}{1\linewidth}{l ccc ccc}
\midrule[2pt]
&Cora&Citeseer&Pubmed &Cora & Citeseer&Pubmed \\
\midrule[2pt]
\rothead{\centering{Trained Dense-16}} 
                        &   \includegraphics[width=0.15\textwidth,valign=m]{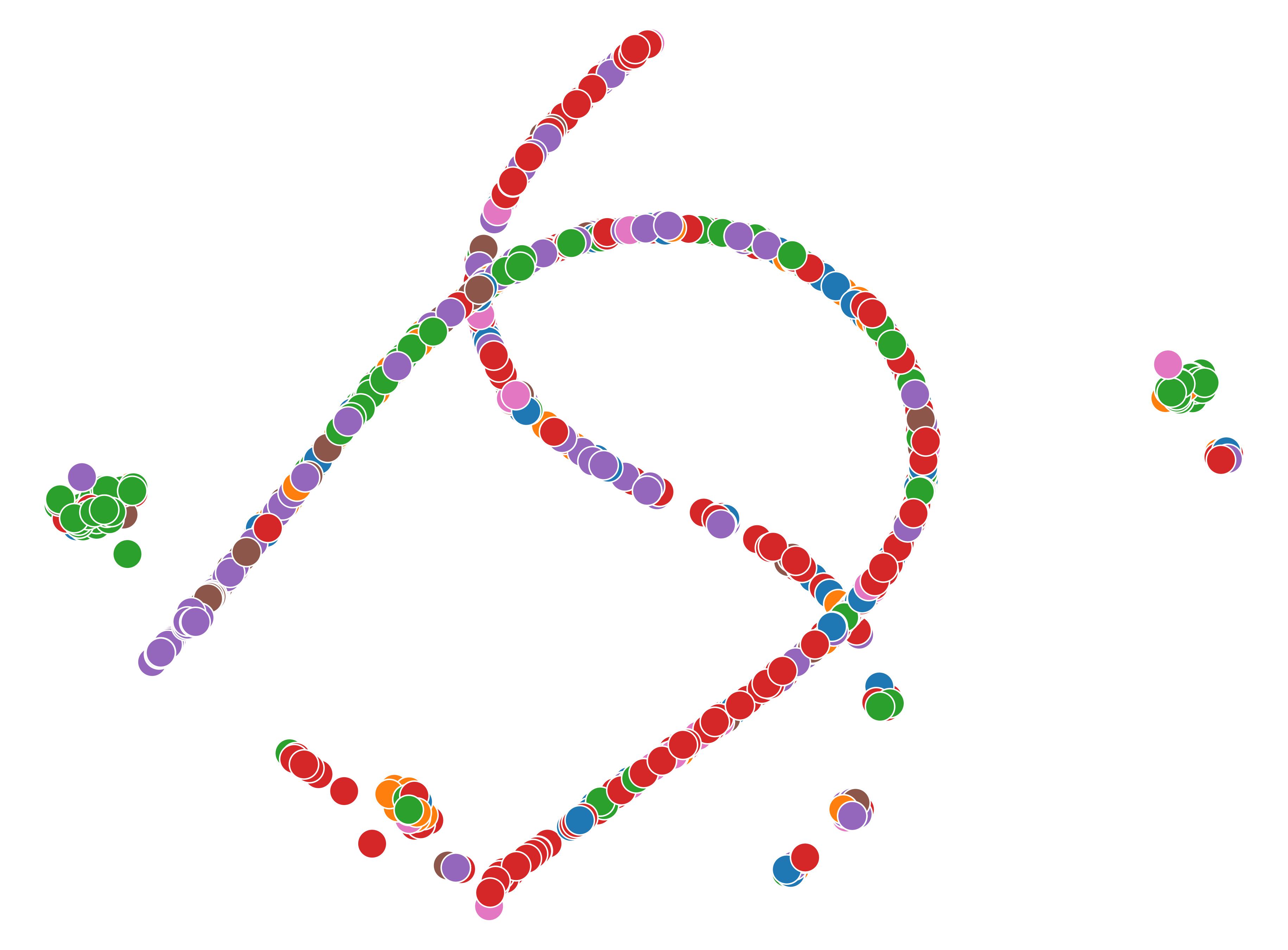}    
                        &   \includegraphics[width=0.15\textwidth,valign=m]{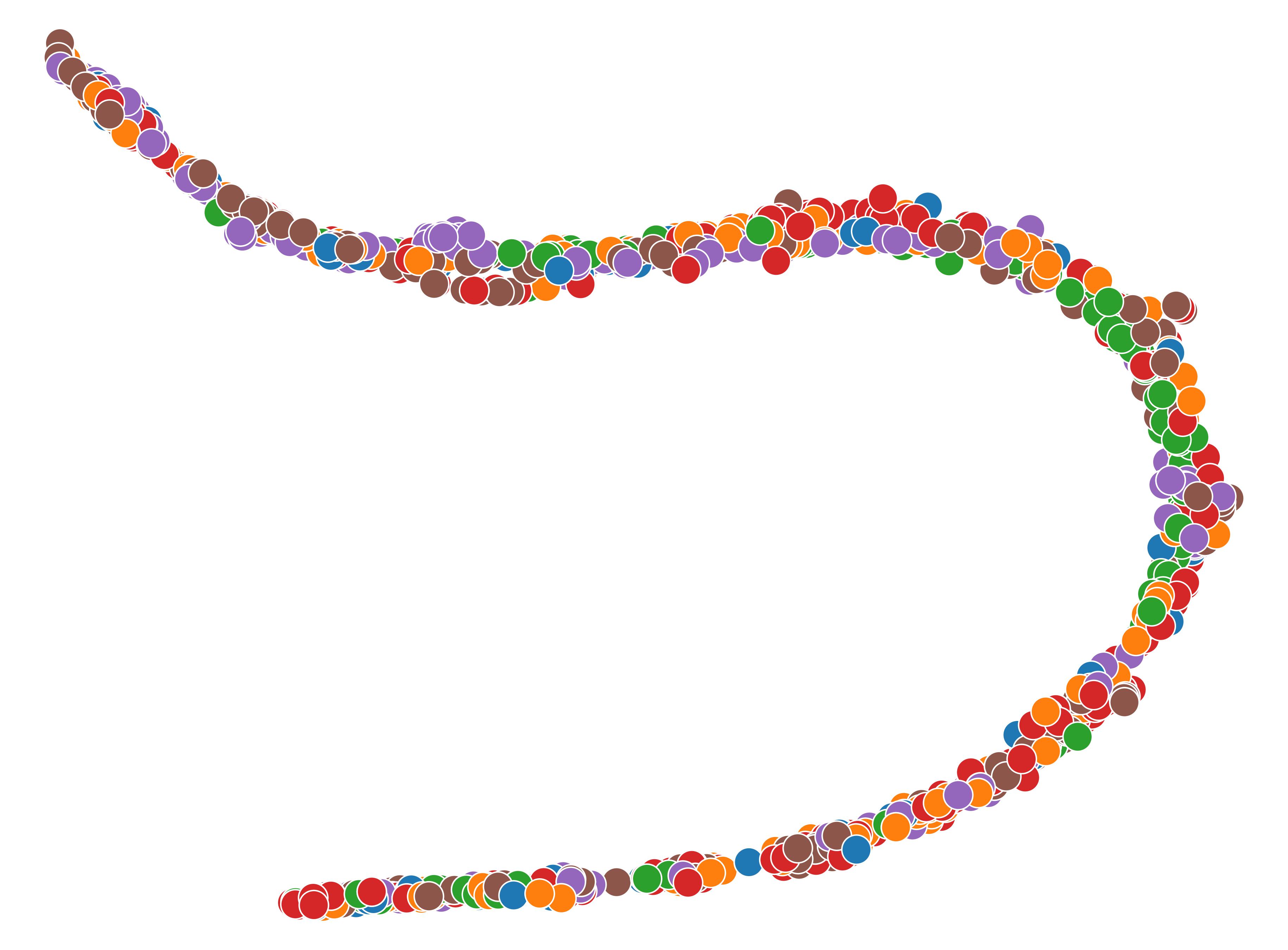}
                        &   \includegraphics[width=0.15\textwidth,valign=m]{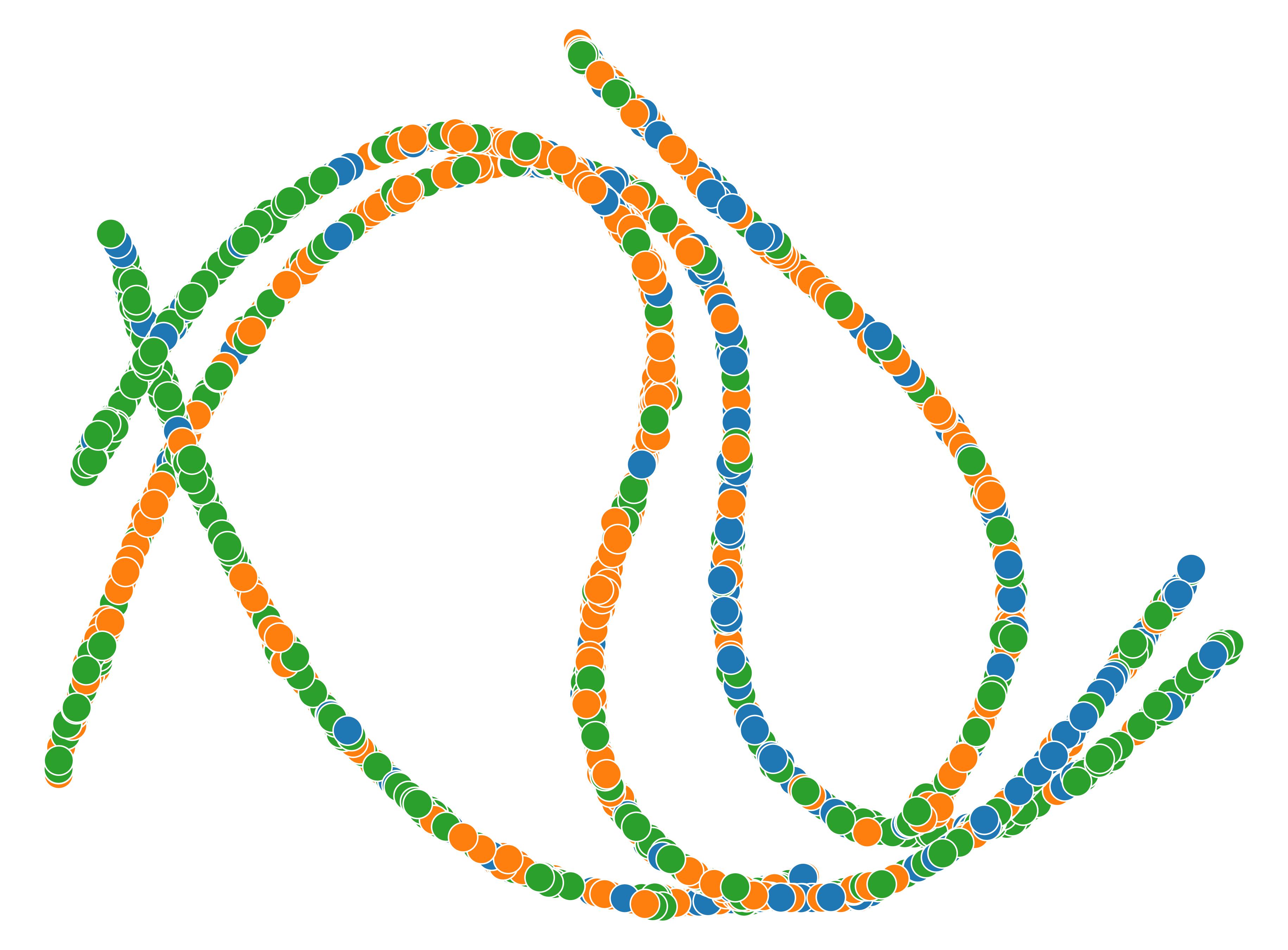}
\rothead{\centering{Trained Dense-32}} 
                        &   \includegraphics[width=0.15\textwidth,valign=m]{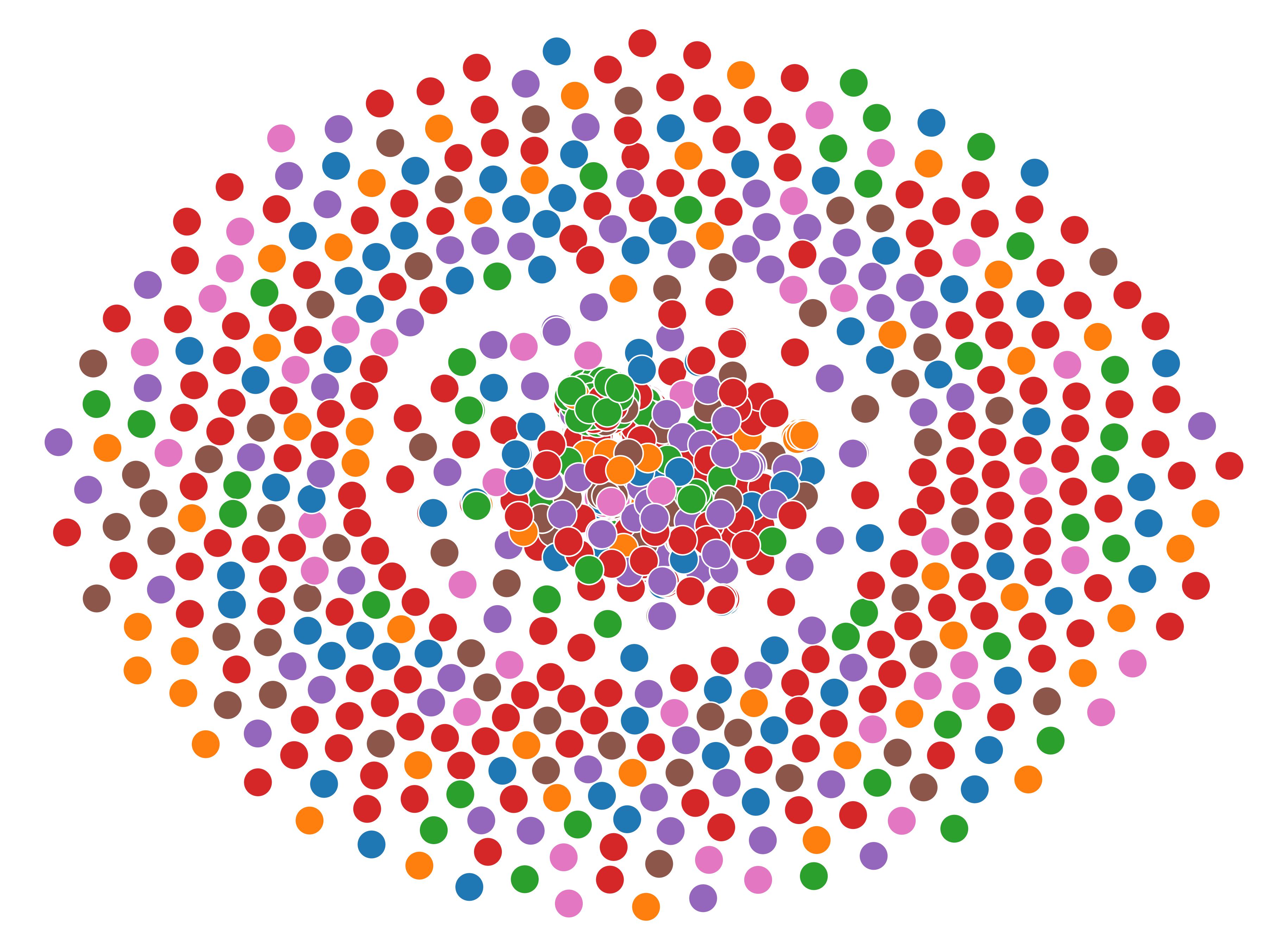}    
                        &   \includegraphics[width=0.15\textwidth,valign=m]{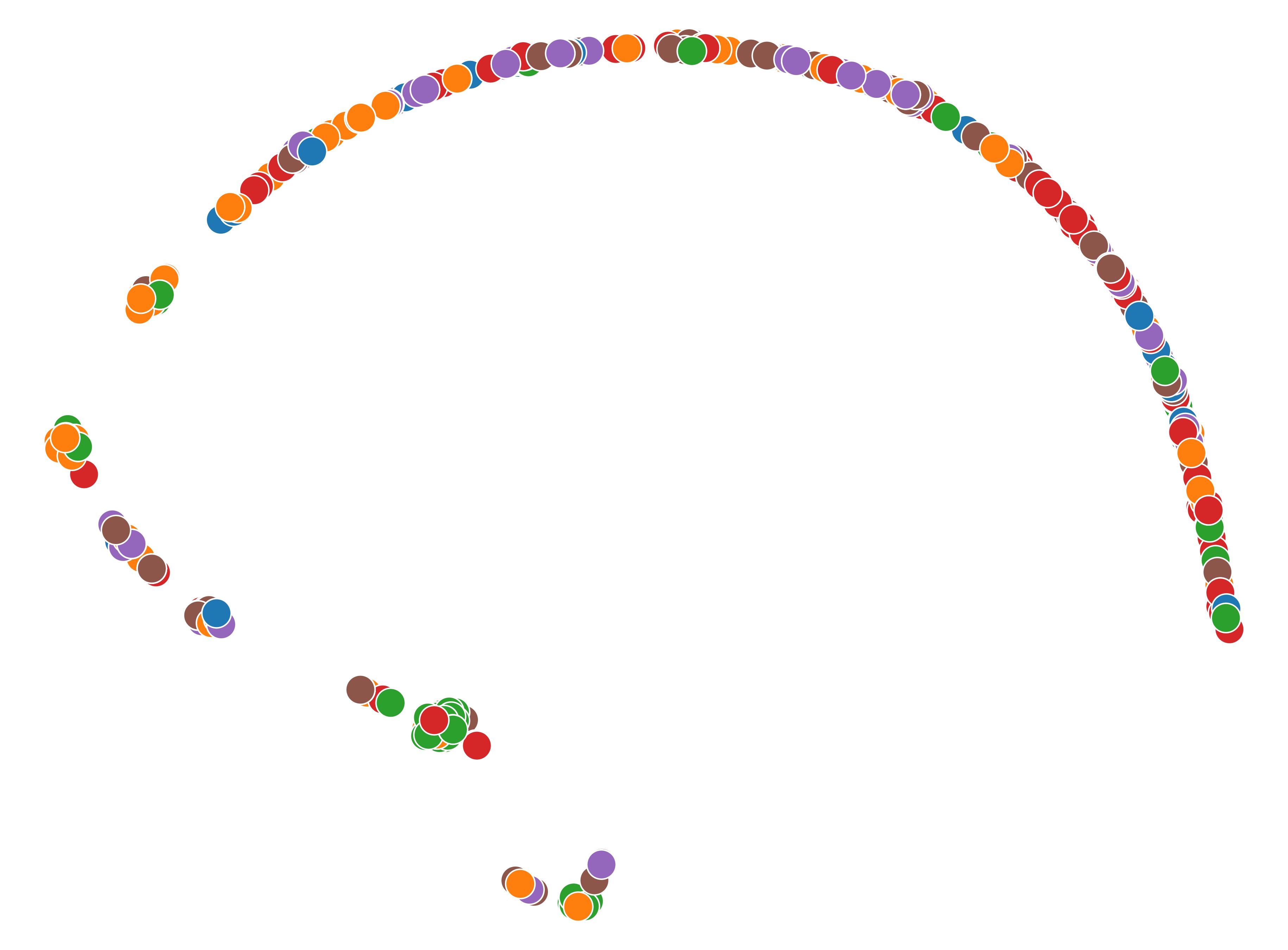}
                        &   \includegraphics[width=0.15\textwidth,valign=m]{Figures/TSNE/Pubmed_GCN_1000_normal_32.jpg}
                        \\[-0.5em]
\midrule[0.1pt]
\rothead{\centering{UGTs-16}} 
                        &   \includegraphics[width=0.15\textwidth,valign=m]{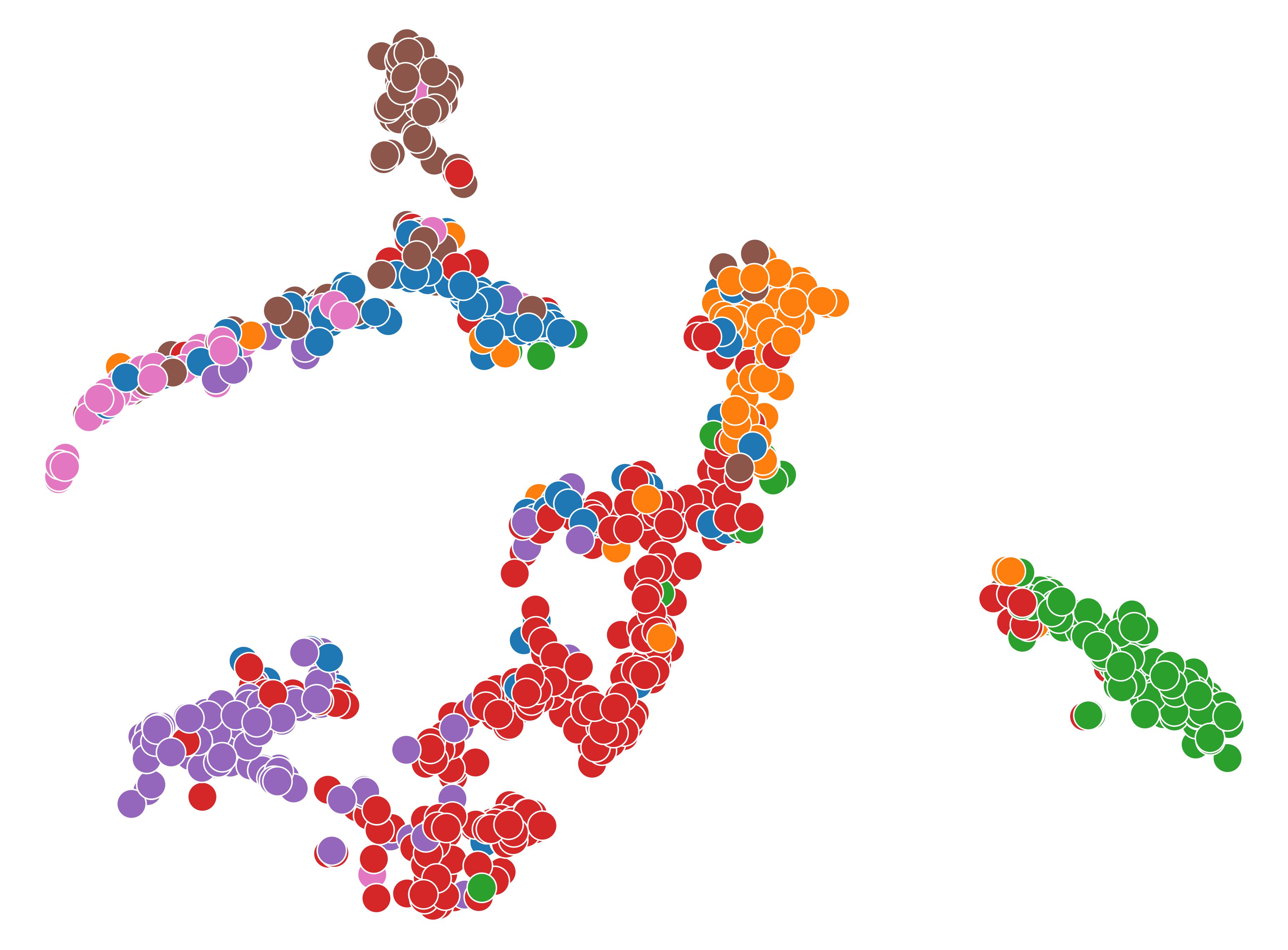}    
                        &   \includegraphics[width=0.15\textwidth,valign=m]{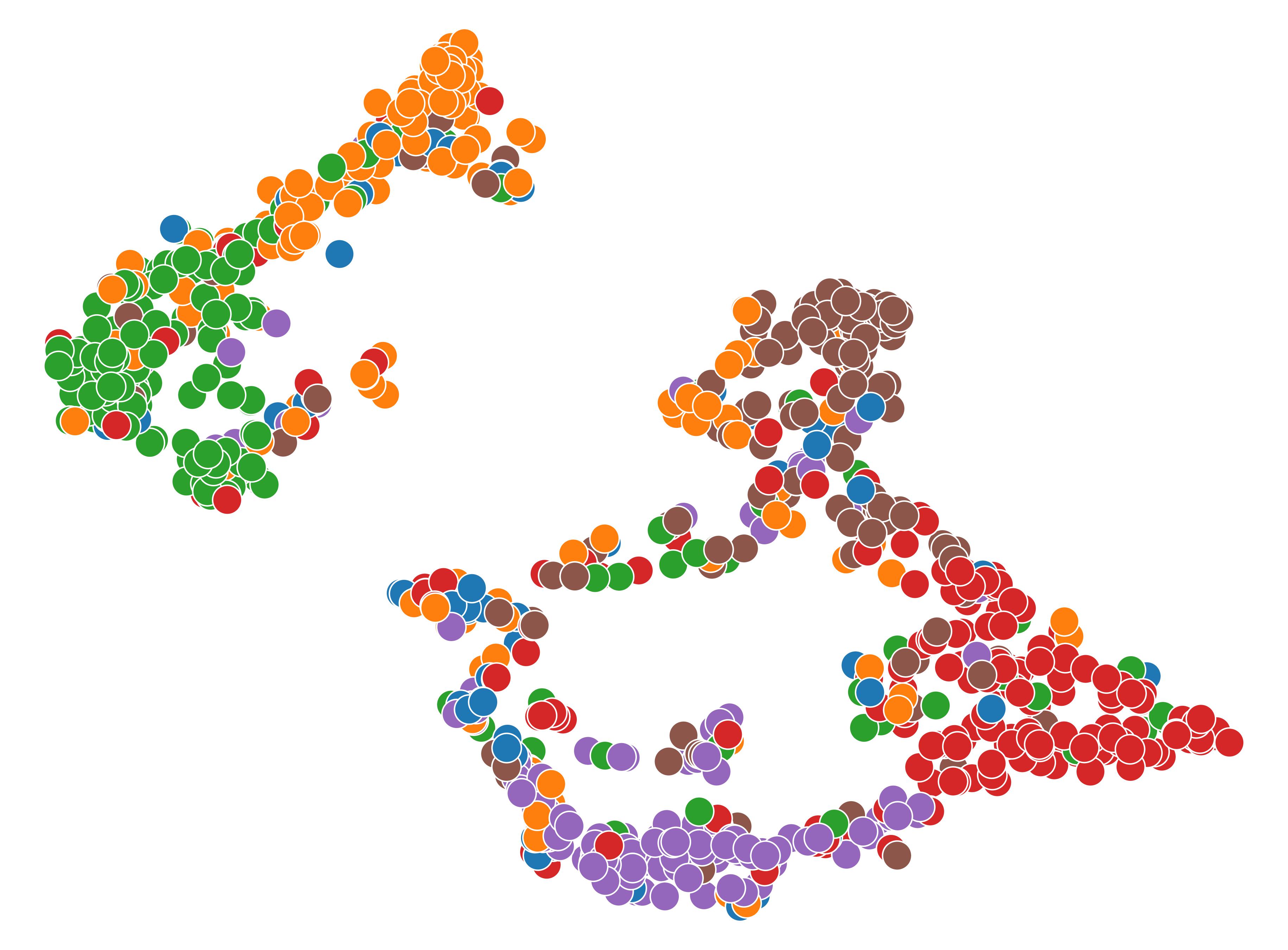}
                        &   \includegraphics[width=0.15\textwidth,valign=m]{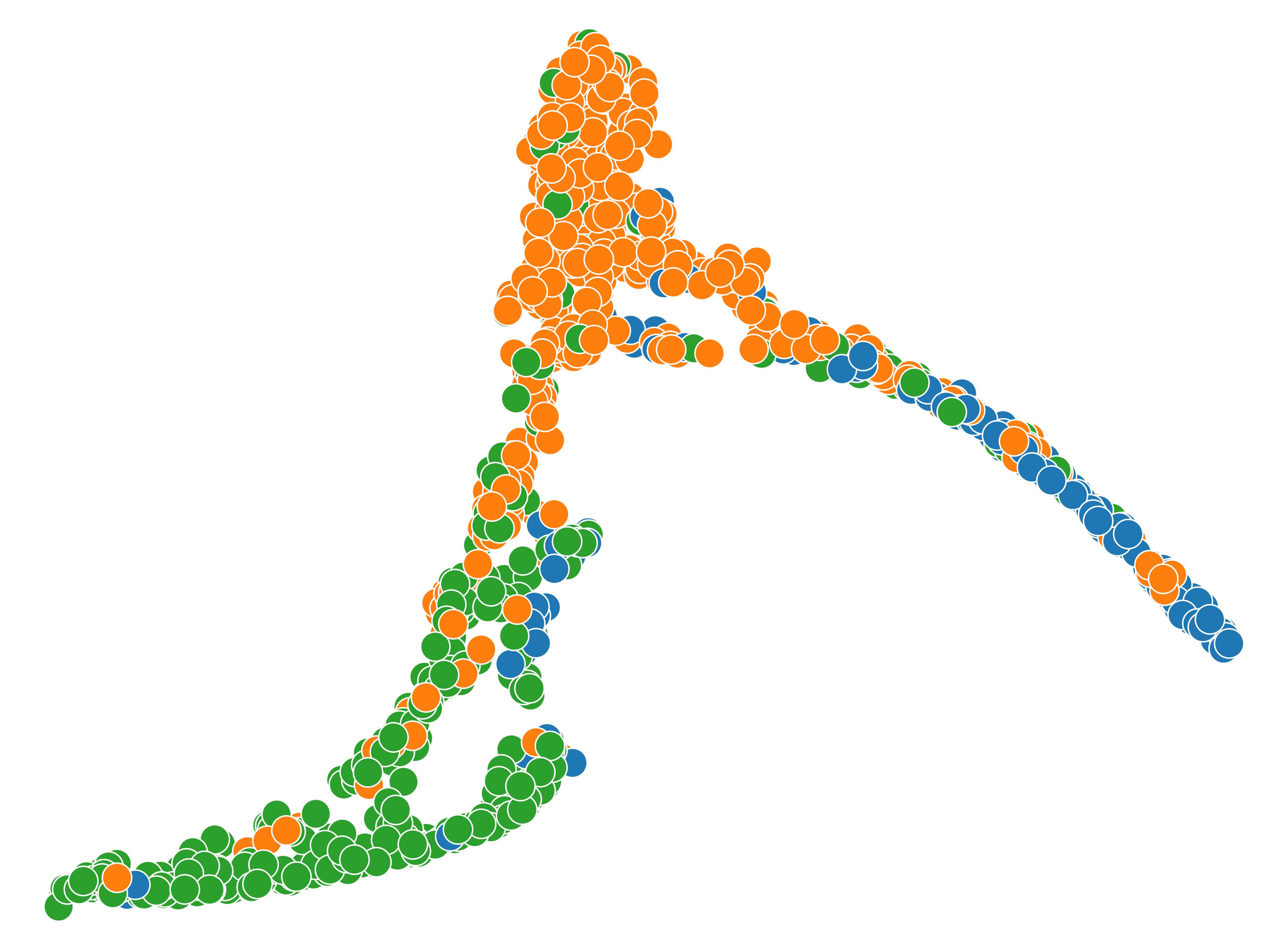}
\rothead{\centering{UGTs-32}} 
                        &   \includegraphics[width=0.15\textwidth,valign=m]{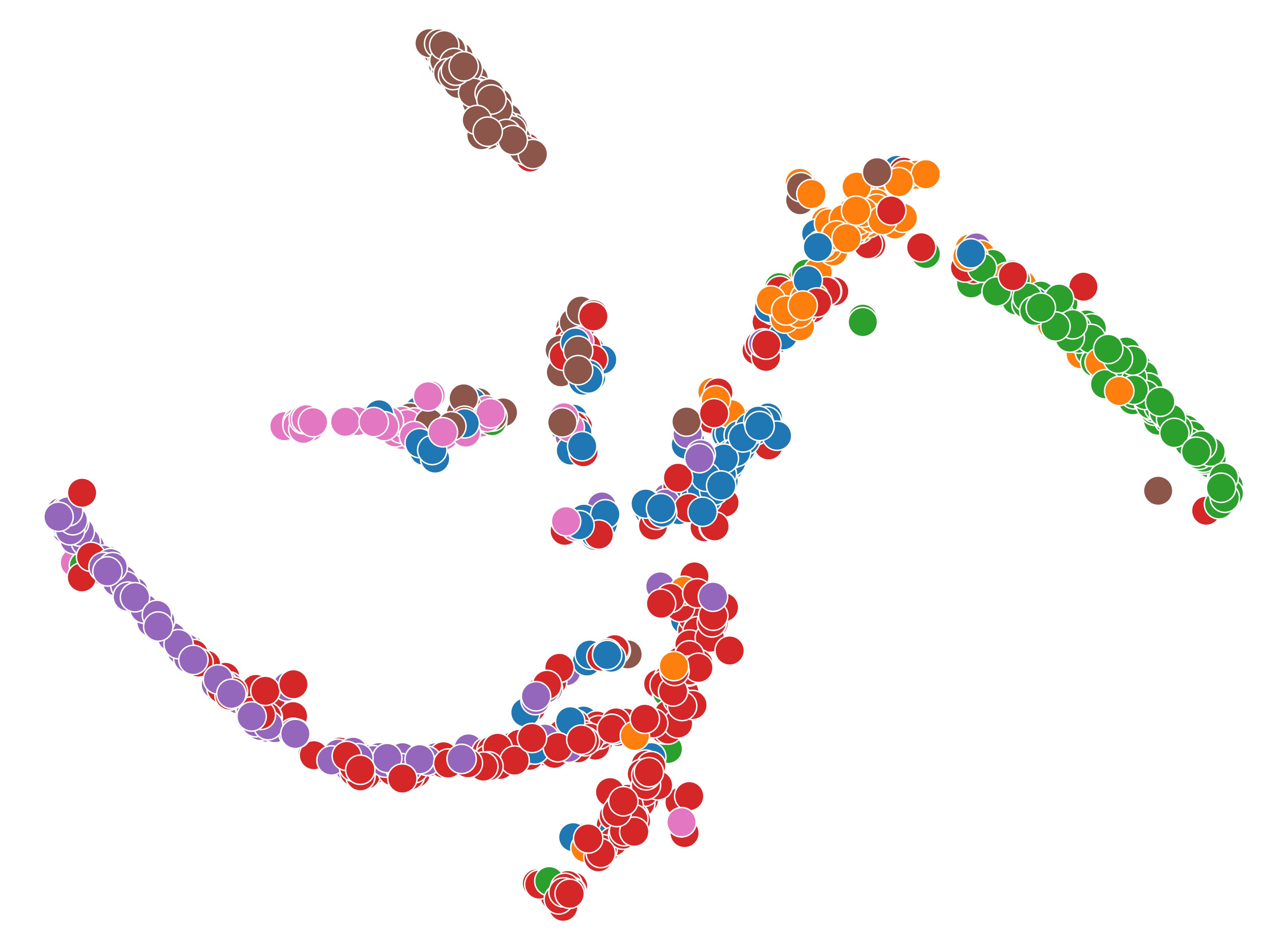}    
                        &   \includegraphics[width=0.15\textwidth,valign=m]{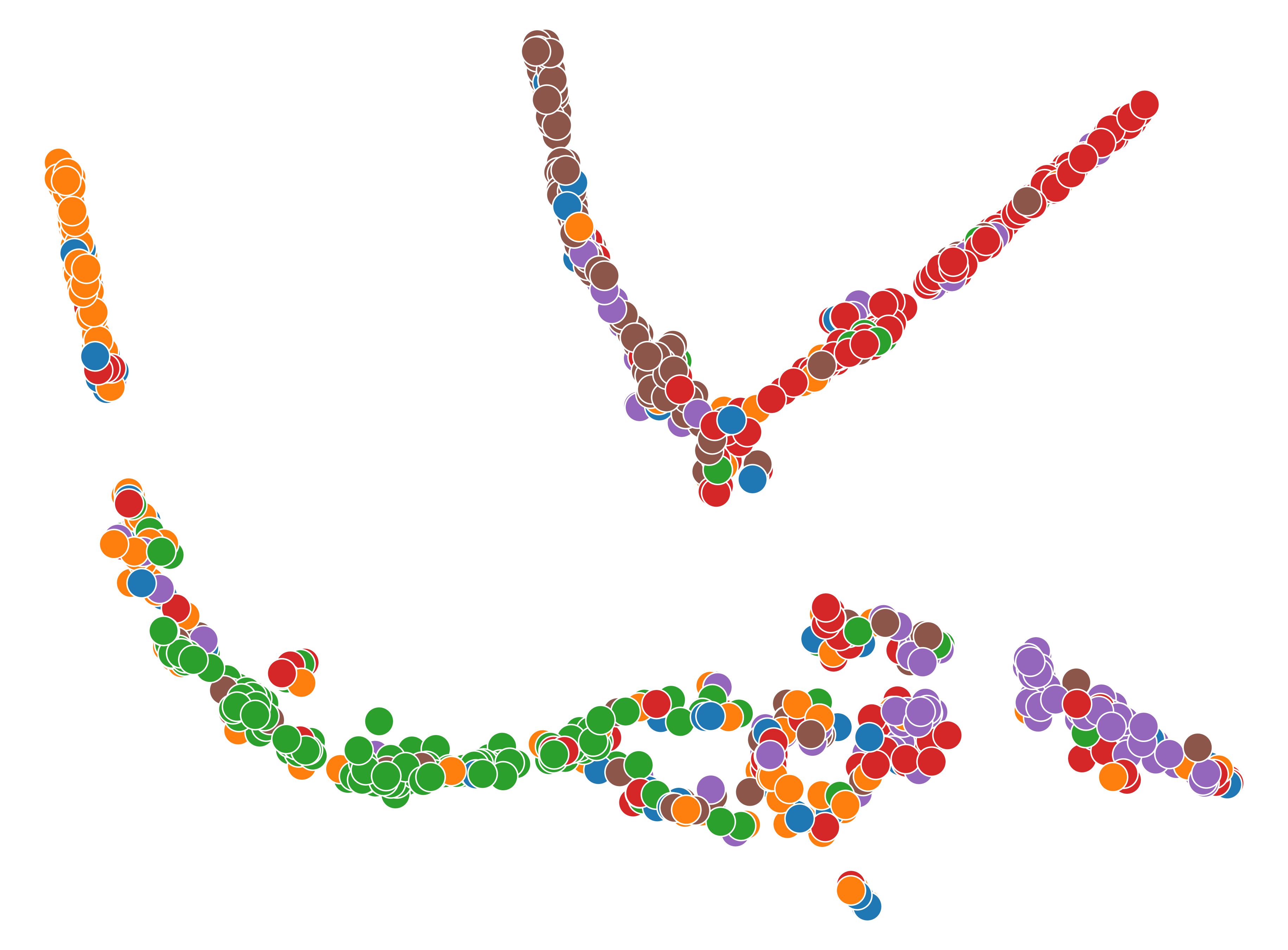}
                        &   \includegraphics[width=0.15\textwidth,valign=m]{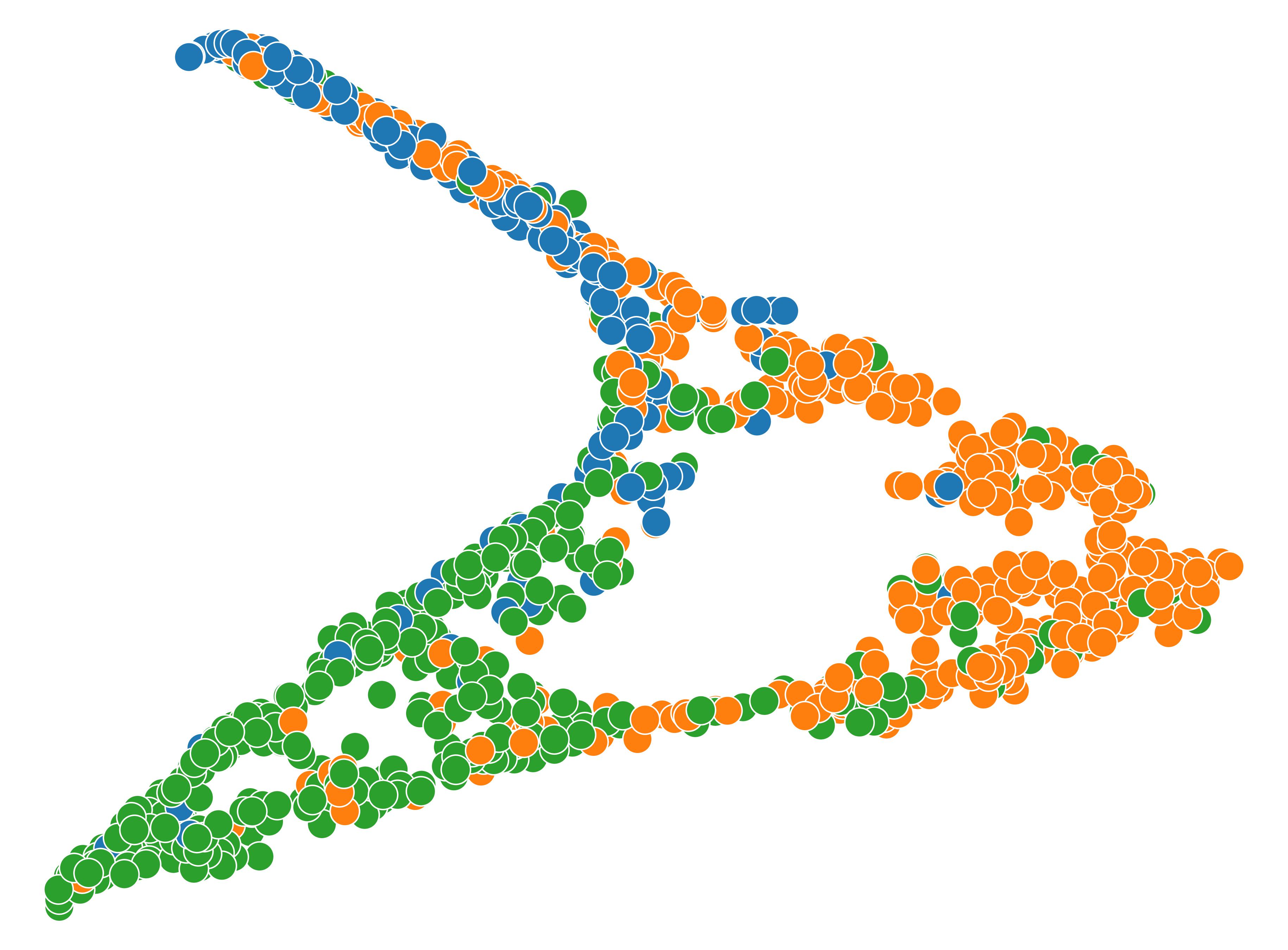} \\
\midrule[2pt]
\end{tabularx}
\end{adjustbox}
\caption{TSNE visualization for node representations. Experiments are based on GAT with fixed width 448.}
\label{Fig:TSNE_GAT}
\end{figure*}

Figure~\ref{Fig:TSNE_GAT} provides the TSNE visualization for node representations learned by UGTs and dense GAT. It can be observed that the node representations learned by the trained dense GAT are mixed while the node representations learned by UGTs are disentangled.

\subsection{Out of distribution detection}\label{appendix_OOD}
\begin{figure*}[!h]
\centering
\subfloat{\includegraphics[width=0.98\textwidth]{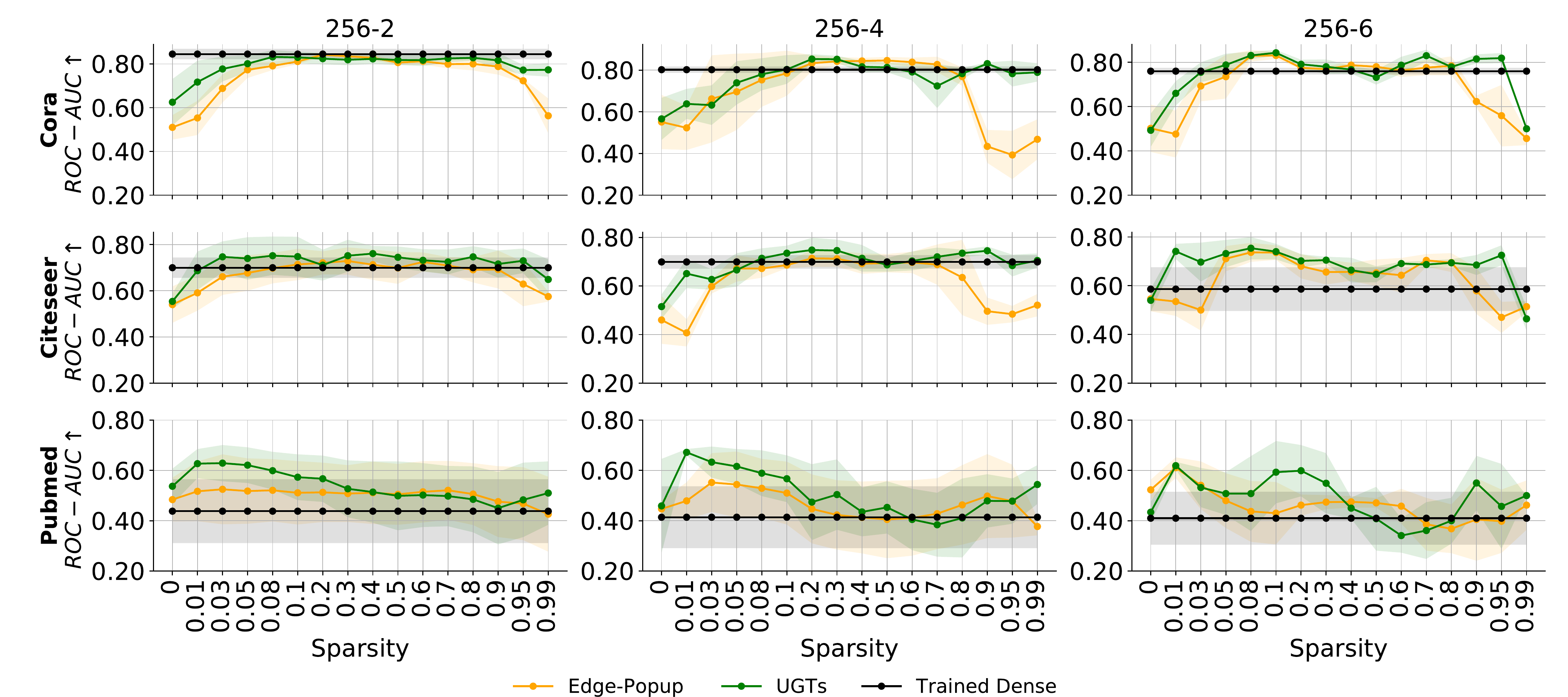}}
\caption{\textbf{Out-of-distribution performance (ROC-AUC)}. Experiments are based on GAT architecture (Width:256, Depth:2)}
\label{fig:OOD_GAT}
\end{figure*}

Figure~\ref{fig:OOD_GAT} shows the OOD performance for UGTs and the trained dense GNNs based on GAT architecture. As we can observe, UGTs achieves very appealing results on OOD performance than the corresponding trained dense GAT.

\subsection{Robustness against input perturbations}
\begin{figure*}[!h]
\centering
\subfloat{\includegraphics[width=0.98\textwidth]{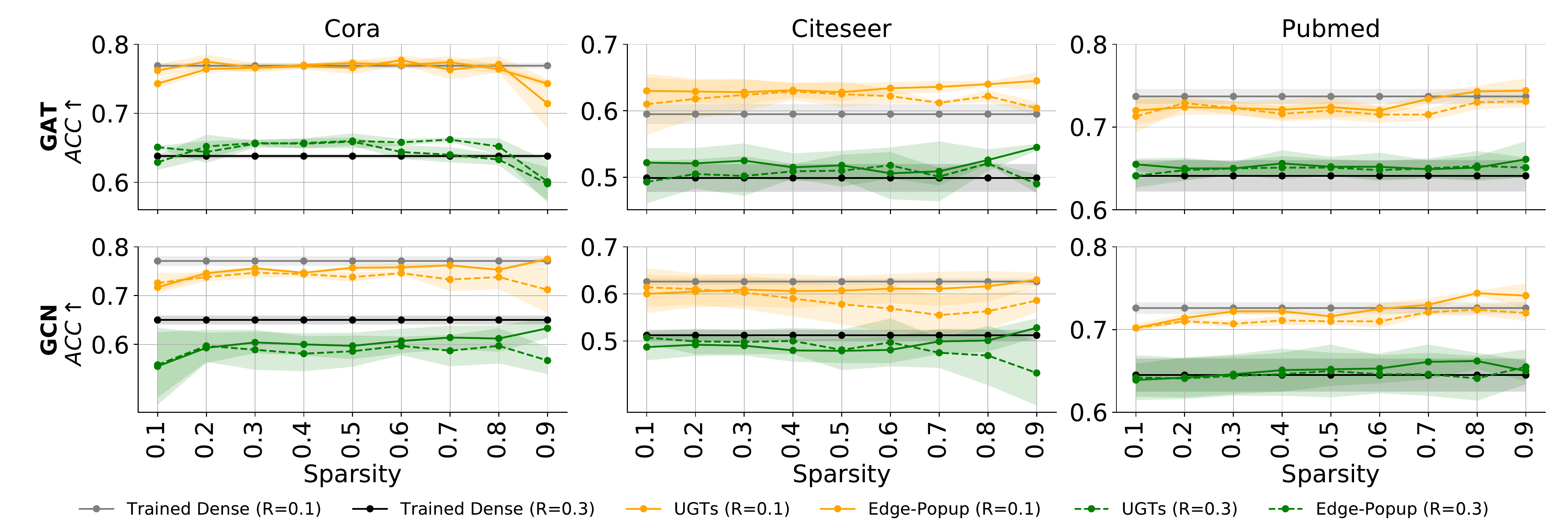}}
\caption{ \textbf{The robust performance on edge perturbations}.$R$ denotes the fraction of perturbed edges.(width:256, Depth:2)}
\label{fig:Robust_Edges}
\end{figure*}

\begin{figure*}[!h]
\centering
\subfloat{\includegraphics[width=0.98\textwidth]{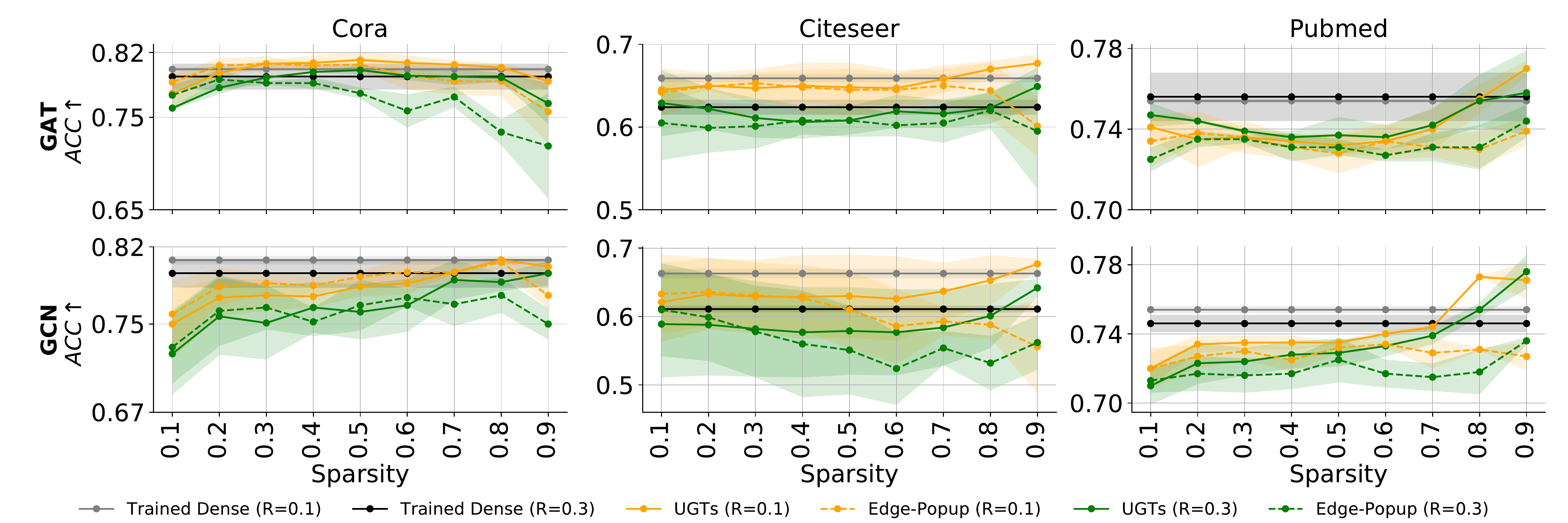}}
\caption{ \textbf{The robust performance on feature perturbations}.$R$ denotes the fraction of perturbed nodes. (width:256, Depth:2)}
\label{fig:Robust_features}
\end{figure*}

In this section, we explore the robustness against input perturbations with varying the sparsity of untrained GNNs. Experiments are conducted on GAT and GCN architectures with width=256 and depth=2. Results are reported in Figure~\ref{fig:Robust_features} and Figure~\ref{fig:Robust_Edges}.

It can be observed that the robustness achieved by UGTs is increasing with the increase of sparsity for both edge and feature perturbation types. Besides, the robustness achieved by UGTs at large sparsity, e.g., sparsity =0.9, can outperform the counterpart trained dense GNNs.

\subsection{The accuracy performance w.r.t model width}

\begin{figure*}[!h]
\centering
\subfloat{\includegraphics[width=0.98\textwidth]{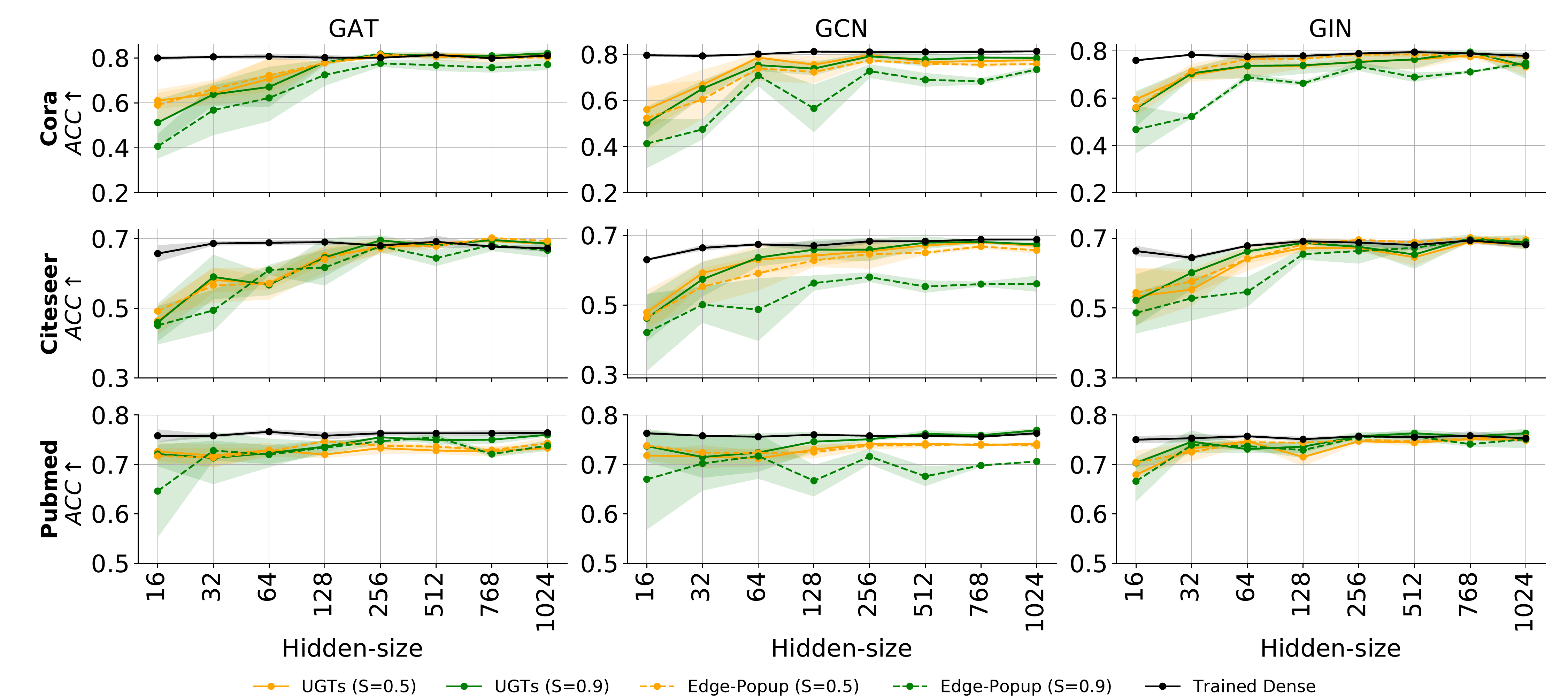}}
\caption{\textbf{Model Width:} The accuracy performance of subnetworks from untrained GNNs w.r.t varying Hidden-Size. The ``S01,S05,S09'' represents the sparsity of the untrained GNNs. The dashed line represents the results of the trained dense GNNs. Experiments are based on GNNs with 2 layers.}
\label{fig:hidden_size}
\end{figure*}

Figure~\ref{fig:hidden_size} shows the performance of UGTs on different architectures with varying model width from 16 to 1024 and fix depth=2. We summarize observations as follows:

\textbf{\circled{1} Performance of UGTs improves with the width of the GNN models.} With width increasing from 16 to 256, the performance of UGTs improves apparently and after width=256, the benefits from model width are saturated.
\subsection{Ablation studies}
We conduct the ablation studies to show the effectiveness of UGTs.The results showed on Table~\ref{tab:ablations}. Compared with Edge-Popup, UGTs mainly has two novelties: global sparsification (VS. uniform sparsification) and gradual sparsification (VS. one-shot sparsification). Here we compare UGTs with 3 baselines: (1) UGTs - global sparsification; (2) UGTs - gradual sparsification; (3) Edge-Popup. The results are reported in the following table and it shows that global sparsification plays an important role for finding  important weights and gradual sparsification is crucial for further boosting performance at high sparsity level. 

\begin{table}[H]
\caption{Ablation studies based on GAT (Depth:2, Width:256) and Cora.}
    \centering
    \begin{tabular}{l|c|c|c|c|c|c}
    \hline
    &0.1&0.3&0.5&0.7&0.9&0.95\\
    \hline
         Edge-Popup&\textbf{0.814}&0.81&0.809&0.81&0.791&0.461\\
         \hline
         UGTs - global sparsification&0.807&0.816&0.817&0.804&0.799&0.731\\
         \hline
         UGTs - gradual sparsification&0.806&\textbf{0.818}&\textbf{0.821}&\textbf{0.821}&0.804&0.795\\
         \hline
         UGTs&0.797&0.811&0.81&0.815&\textbf{0.817}&\textbf{0.822}\\
         \hline
    \end{tabular}
    \label{tab:ablations}
\end{table}

\subsection{Observations via gradient norm}
\begin{figure}[htb]
\centering
\subfloat[Epoch=10]{\includegraphics[width=0.35\textwidth]{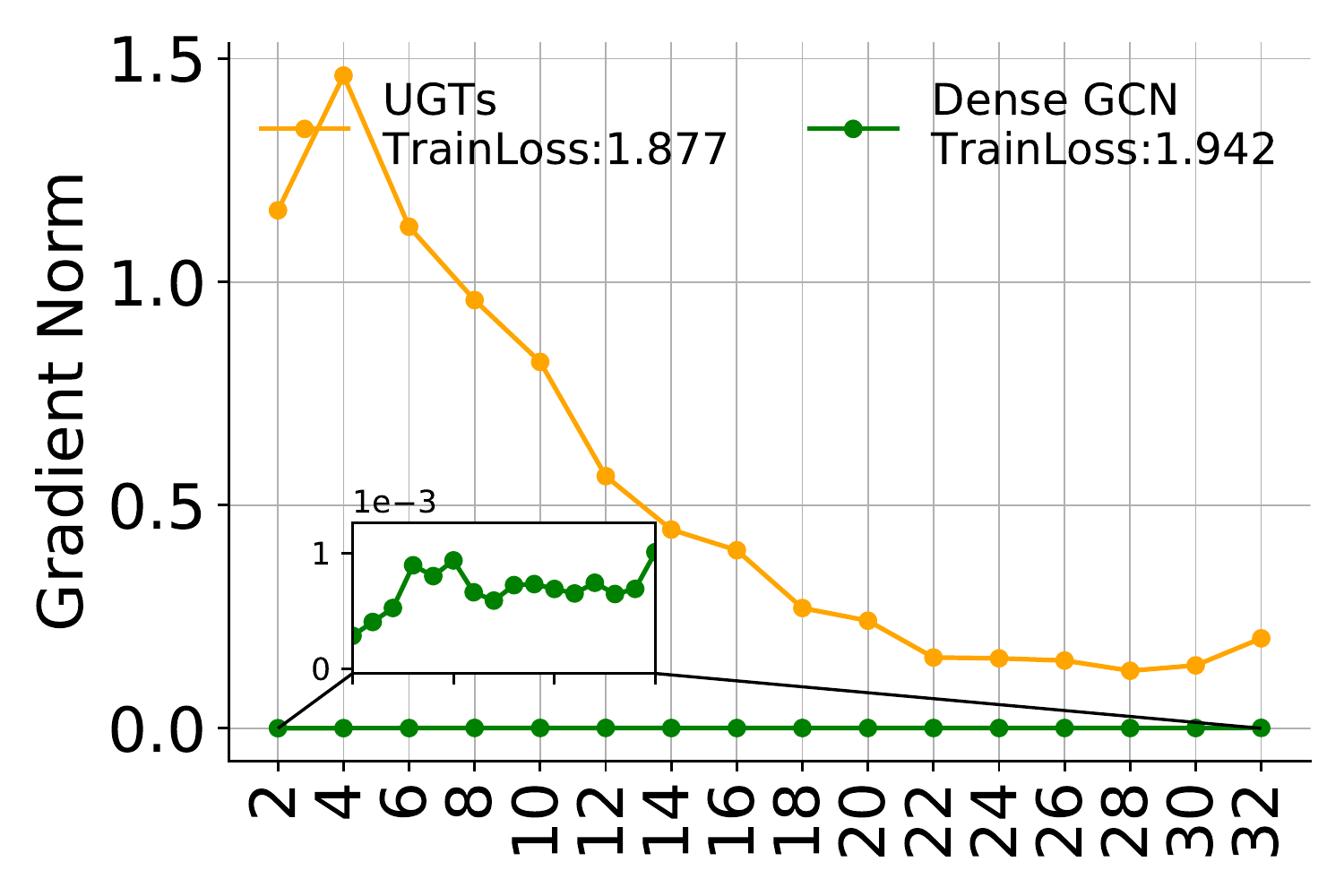} 
}
\subfloat[Epoch=150]{\includegraphics[width=0.35\textwidth]{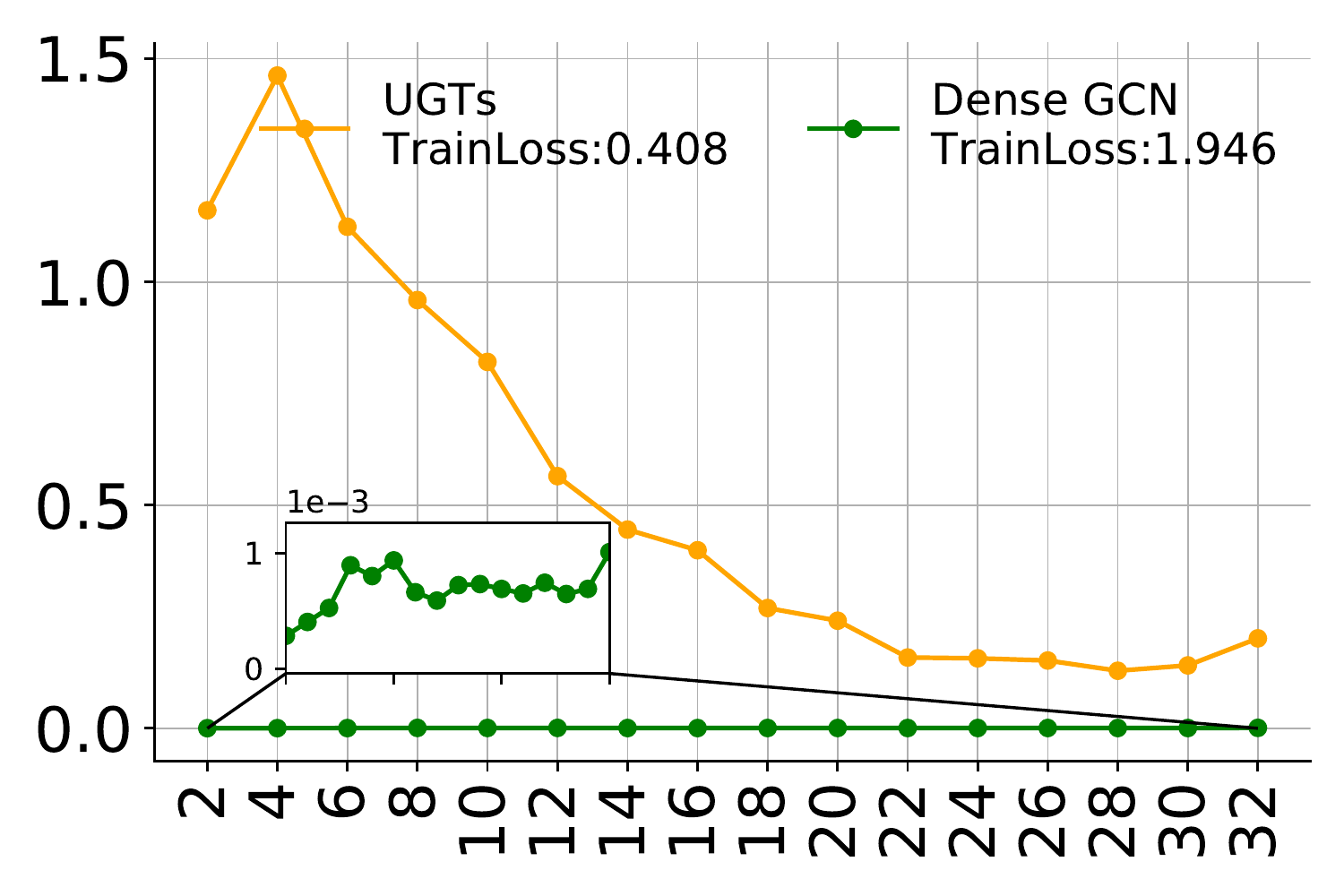} 
}\\
\subfloat[Epoch=270]{\includegraphics[width=0.35\textwidth]{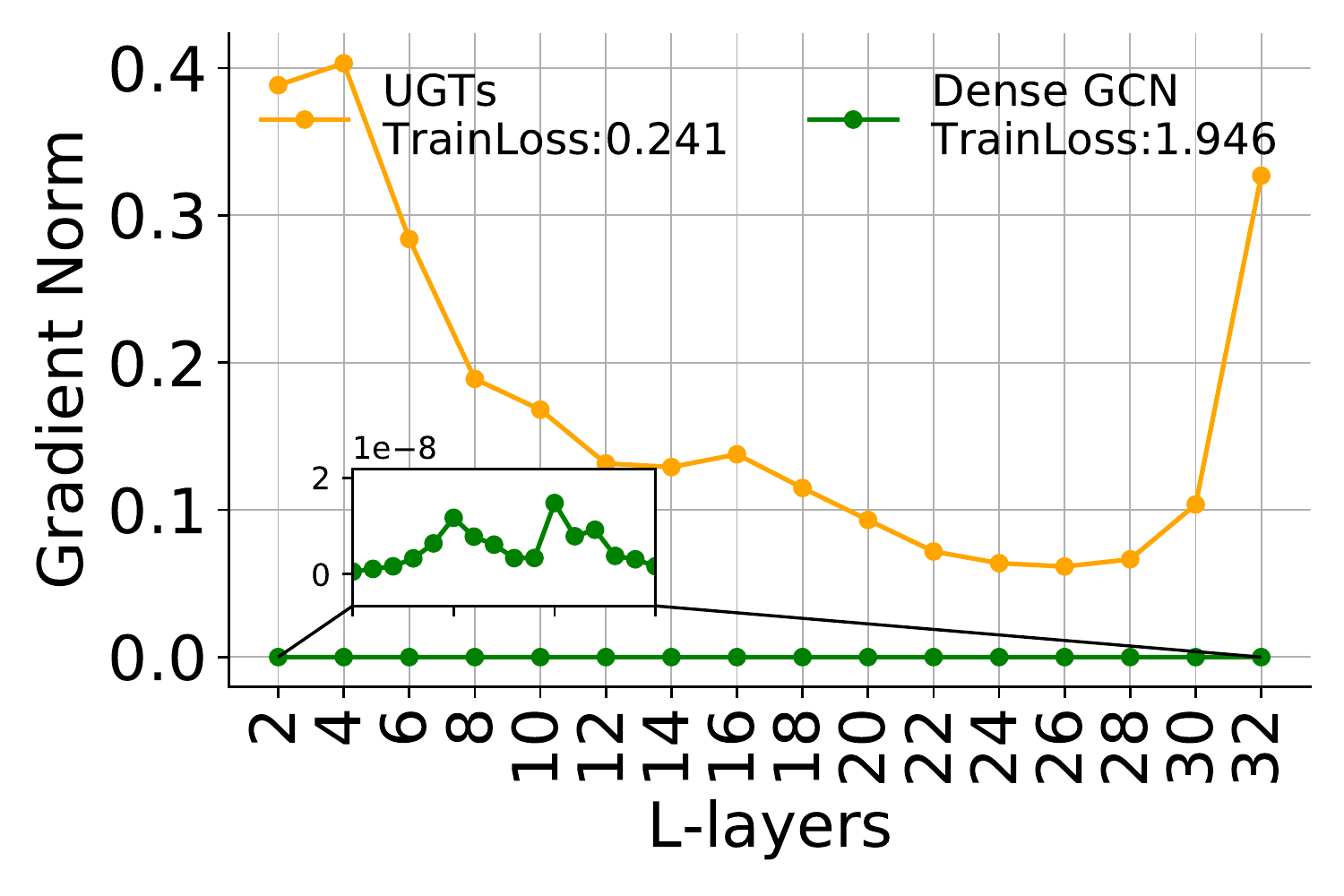} 
}
\subfloat[Epoch=400]{\includegraphics[width=0.35\textwidth]{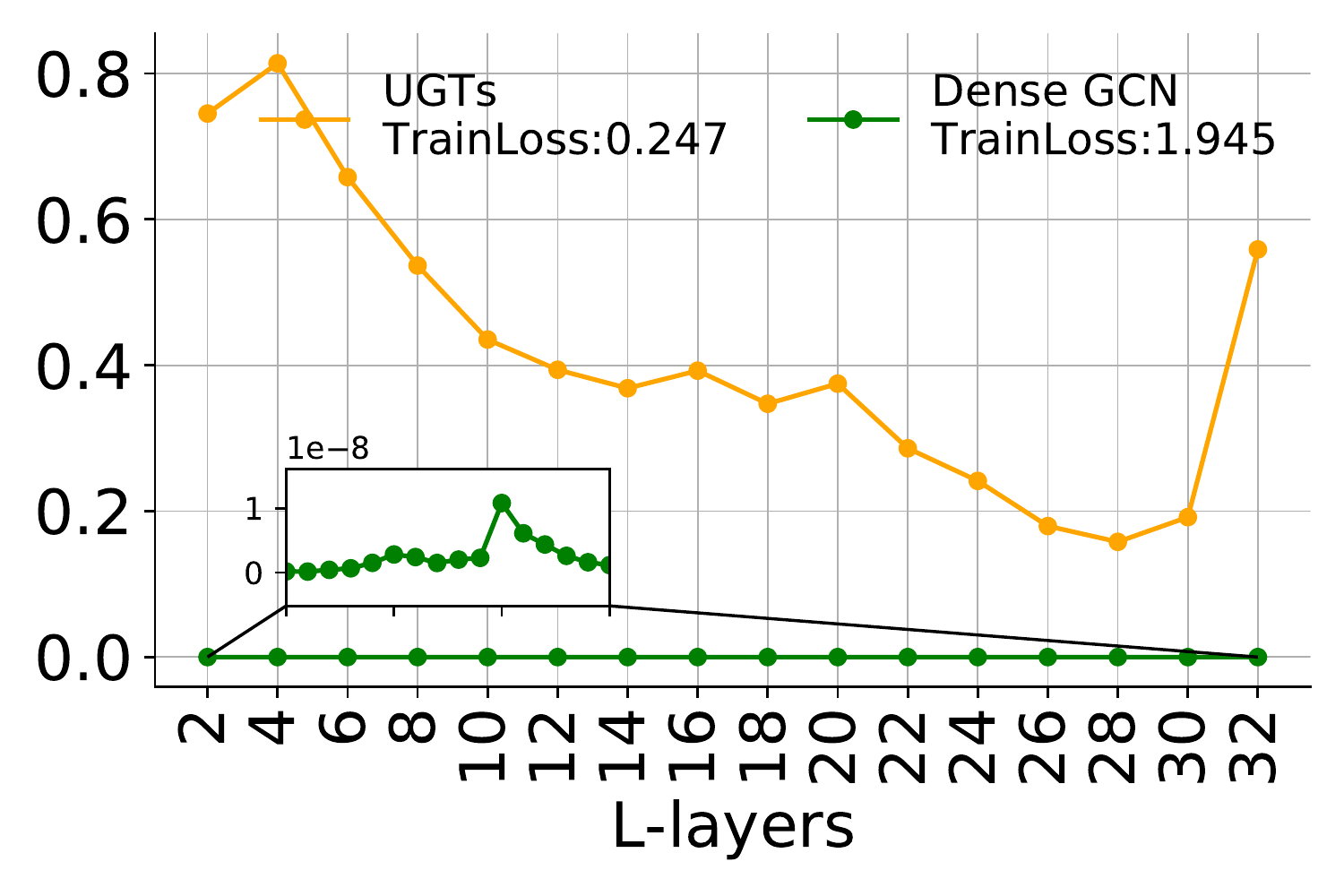} 
}
\caption{\textbf{Gradient norm w.r.t each layer during training.} Experiments are conducted on Cora with GCN architecture containing 32 layers and width 448.}
\label{fig:gradientnorm}
\end{figure}

To preliminary understand why UGTs can mitigate over-smoothing while the trained dense GNNs can not, we calculate the gradient norm of each layer for UGTs and dense GCN during training. In order to have a fair comparison, we calculate the gradient norm of $\nabla_{(\bm{m}^{l}\odot \bm{\theta}^{l})} \mathcal{L}(g(\bm{A},\bm{X};\bm{\theta} \odot \bm{m}),y)$ for UGTs and the gradient norm of $\nabla_{\bm{\theta}^{l}} \mathcal{L}(g(\bm{A},\bm{X};\bm{\theta}),y)$ for dense GCN where $l$ denotes the layer. Results are reported in Figure~\ref{fig:gradientnorm}.

As we can observe, the gradient vanishing problem may exist for training deep dense GCN since the gradient norm for dense GCN is extremely small while UGTs does not have this problem. This problem might also be indicated by the training loss where the training loss for dense GCN does not decrease while the training loss for UGTs decreases a lot. This might explain why UGTs performs well for deep GNNs.  

\subsection{More experiments for mitigating over-smoothing problem}\label{appedix_oversmoothing}
We conduct experiments on Cora, Citeseer and Pubmed for GAT with deeper layers. The width is fixed to 448. The results of the other methods are obtained by running the code\footnote{\url{https://github.com/VITA-Group/Deep_GCN_Benchmarking.git}}.

The results are reported in Table~\ref{tab:effective_oversmooth_GAT}. It can be observed again that UGTs consistently outperforms all the baselines. 

\begin{table*}[htb]
\centering
\caption{Test accuracy (\%) of different training techniques. The experiments are based on GAT models with 16, 32 layers, respectively. Width is set to 448.}
\begin{adjustbox}{width=0.8\textwidth}
\begin{tabular}{c |cc|cc|cc}
\toprule[2pt]
   &\multicolumn{2}{c|}{Cora} &\multicolumn{2}{c|}{Citeseer}&\multicolumn{2}{c}{Pubmed}\\
\midrule[1pt]
    N-Layers&   16 & 32 &   16 & 32 &  16 & 32   \\
\midrule[1.5pt]
    \textbf{Trained Dense GAT}  &20.6 &13.0&20.0&16.9&17.9&18.0\\
\midrule[1pt]
    +Residual  & 19.9& 20.7&17.7&19.2&41.6&40.8  \\
     +Jumping  & 39.7& 27.8&29.1&25.5&57.3&57.1 \\
\midrule[1pt]
    +NodeNorm &70.9&11.0&17.1&18.4&72.2&59.7\\
    +PairNorm &27.9&12.1&22.8&17.7&73.0&44.0\\
\midrule[1pt]
    +DropNode &23.6&13.0&18.8&7.0&26.7&18.0 \\
    +DropEdge &24.8&13.0&19.4&7.0&19.3&18.0\\
\midrule[2pt]
    \textbf{UGTs-GAT} & \textbf{76.7 $\pm$ 1.1} &\textbf{74.9$\pm$0.2}  &\textbf{62.7$\pm$0.7} &\textbf{56.5$\pm$1.1} &\textbf{77.9$\pm$0.5} &\textbf{75.5$\pm$1.5}\\   
\bottomrule[2pt]
\end{tabular}
\end{adjustbox}
\label{tab:effective_oversmooth_GAT}
\end{table*}

\section{Pseudocode}\label{apped_psu}

Pseudocode is showed in Algothrim~\ref{alg:UGTs}.

\begin{algorithm}[!t]
  \caption{Untrained GNNs Tickets (UGTs)}
  \label{alg:UGTs}
\begin{algorithmic}
  \STATE {\bfseries Input:} a GNN $g(\bm{A},\bm{X};\bm{\theta})$, initial mask $\bm{m}=1 \in \mathcal{R}^{|\bm{\theta}|}$ with latent scores $\bm{S}$, learning rate $\lambda$, hyperparameters for the gradual sparsification schedule $s_i$, $s_f$, $t_0$, and $\Delta t$. 
  \STATE {\bfseries Output:} $g(\bm{A},\bm{X};\bm{\theta} \odot \bm{m}),\bm{y}$ 
  \STATE Randomly initialize model weights $\bm{\theta} $ and $\bm{S}$.
  \FOR{$t=1$ {\bfseries to} $T$}
      \STATE \textcolor{gray}{\#Calculate the current sparsity level $s_{t}$ by Eq.~\ref{gradual_decay}.}
      \STATE $s_{t} \longleftarrow  s_{f}+(s_{i}-s_{f})(1-\frac{t-t_{0}}{n \Delta t})$ 
      \STATE \#Get the global threshold value at top $s_{t}$ by sorting $\bm{S}$ in ascending order.
      \STATE $\bm{S}_{thres} \longleftarrow  $ $Thresholding(\bm{S}, s_{t})$
      \STATE \textcolor{gray}{\#Generate the binary mask.}
      \STATE $\bm{m} \longleftarrow 0$ if $\bm{S}< \bm{S}_{thres}$ else  $1$
      \STATE \textcolor{gray}{\#Update $\bm{S}$.}
      \STATE $\bm{S} \longleftarrow \bm{S}- \lambda \nabla_{\bm{S}}\mathcal{L} (g(A,\bm{X};\bm{\theta} \odot \bm{m}),\bm{y})$
  \ENDFOR
  \STATE Return $g(\bm{A},\bm{X};\bm{\theta} \odot \bm{m}),\bm{y}$
\end{algorithmic}
\end{algorithm}

\end{document}